\RequirePackage[dvipsnames, table]{xcolor}
\documentclass{article}
\usepackage{PRIMEarxiv}

\usepackage{transparent}
\usepackage{graphicx}
\usepackage{multirow}
\usepackage{amsmath,amssymb,amsfonts}
\usepackage{amsthm}
\usepackage{siunitx}
\usepackage{physics}
\usepackage{mathrsfs}
\usepackage{textcomp}
\usepackage{manyfoot}
\usepackage{booktabs}
\usepackage{algorithm}
\usepackage{algorithmicx}
\usepackage{algpseudocode}
\usepackage{listings}
\usepackage{cancel}
\usepackage[normalem]{ulem}
\usepackage{soul}
\usepackage[figuresright]{rotating}
\usepackage[utf8]{inputenc} 
\usepackage[T1]{fontenc}    
\usepackage{hyperref}       
\usepackage{url}            
\usepackage{nicefrac}       
\usepackage{microtype}      
\usepackage{fancyhdr}       
\usepackage{pifont}

\makeatletter

\definecolor{dark21}{HTML}{1b9e77}
\definecolor{dark22}{HTML}{d95f02}
\definecolor{dark23}{HTML}{7570b3}
\definecolor{dark24}{HTML}{e7298a}
\definecolor{dark25}{HTML}{66a61e}
\definecolor{dark26}{HTML}{e6ab02}
\definecolor{dark27}{HTML}{a6761d}
\definecolor{dark28}{HTML}{666666}

\DeclareMathOperator*{\argmin}{arg\,min}
\DeclareMathAlphabet\mathbfcal{OMS}{cmsy}{b}{n}

\newcommand{\xmark}{\ding{54}}

\usepackage{algorithm, algorithmicx, algpseudocode}
\newcommand*{\algrule}[1][\algorithmicindent]{\hspace*{.5em}\vrule\vrule width 0pt height \baselineskip depth .25\baselineskip\hspace*{\dimexpr#1-.5em}}
\makeatletter
\newcount\ALG@printindent@tempcnta
\def\ALG@printindent{%
    \ifnum \theALG@nested>0
    \ifx\ALG@text\ALG@x@notext
    \else
    \unskip
    \ALG@printindent@tempcnta=1
    \loop
    \algrule[\csname ALG@ind@\the\ALG@printindent@tempcnta\endcsname]%
    \advance \ALG@printindent@tempcnta 1
    \ifnum \ALG@printindent@tempcnta<\numexpr\theALG@nested+1\relax
    \repeat
    \fi
    \fi
}%
\usepackage{etoolbox}
\patchcmd{\ALG@doentity}{\noindent\hskip\ALG@tlm}{\ALG@printindent}{}{\errmessage{failed to patch}}
\makeatother
\AtBeginEnvironment{algorithmic}{\lineskip0pt}
\usepackage[scaled]{beramono}
\usepackage[T1]{fontenc}
\algtext*{EndWhile}
\algtext*{EndFor}
\algtext*{EndIf}
\algdef{SE}[DOWHILE]{Do}{doWhile}{\algorithmicdo}[1]{\algorithmicwhile\ #1}%
\algnewcommand\algorithmicto{\textbf{to}}
\algrenewtext{While}{\textbf{while}\ }
\algrenewtext{Procedure}[2]{\textbf{function}\ \textproc{#1}\ifthenelse{\equal{#2}{}}{}{(#2)}}

\newcommand*\Let[2]{\State #1 $\gets$ #2}
\algrenewcommand\algorithmicrequire{\textbf{Input:}}
\algrenewcommand\algorithmicensure{\textbf{Output:}}

\usepackage{eqparbox}
\algrenewcommand{\algorithmiccomment}[1]{\hfill\eqparbox{COMMENT}{\color{gray} \it-- #1}}
\newcommand{\inlinecomment}[1]{{\color{gray} \it-- #1}}

\title{Neural topology optimization: the good, the bad, and the ugly}

\author{
  Suryanarayanan Manoj Sanu, Alejandro M.~Arag\'{o}n \\
  Faculty of Mechanical Engineering\\
  Delft university of technology \\
  Delft\\
  The Netherlands \\
  \texttt{\{s.manojsanu, a.m.aragon\}@tudelft.nl} \\
   \And
  Miguel A.~Bessa \\
School of engineering \\
Brown University\\
Providence\\
United States of America \\
\texttt{miguel\_bessa@brown.edu} \\
}

\begin{document}
\maketitle

\begin{abstract}
Neural networks (NNs) hold great promise for advancing inverse design via topology optimization (TO), yet misconceptions about their application persist. This article focuses on neural topology optimization (neural TO), which leverages NNs to reparameterize the decision space and reshape the optimization landscape. While the method is still in its infancy, our analysis tools reveal critical insights into the NNs' impact on the optimization process. We demonstrate that the choice of NN architecture significantly influences the objective landscape and the optimizer's path to an optimum. Notably, NNs introduce non-convexities even in otherwise convex landscapes, potentially delaying convergence in convex problems but enhancing exploration for non-convex problems. This analysis lays the groundwork for future advancements by highlighting: 1) the potential of neural TO for non-convex problems and dedicated GPU hardware (the ``good''), 2) the limitations in smooth landscapes (the ``bad''), and 3) the complex challenge of selecting optimal NN architectures and hyperparameters for superior performance (the ``ugly'').
\end{abstract}


Nonlinear transformations or reparameterizations have been a central tenet in the success of machine learning. An elementary example is the ``kernel trick'' where data inputs (features) are transformed into higher-dimensional representations that can make the outputs linearly separable in that space \cite{murphy2012a}. This facilitates the training of simpler classification and regression models. More powerful transformations can be obtained by artificial neural networks (NNs), but in essence, the idea remains the same: transform the features, reshape the loss landscape, and improve function representation (learning). The same idea can be applied to optimization. Any objective landscape\footnote{We refer to ``objective landscape'' when the goal is to optimize the objective function (optimization), and ``loss landscape'' when the goal is to minimize error (learning).} can be transformed such that the optimizer can traverse that landscape more effectively. Intuitively, however, not every reparameterization is created equal. Some reparameterizations can make optimization more difficult, while others can make it easier.

\begin{figure*}[t]
\centering
\includegraphics{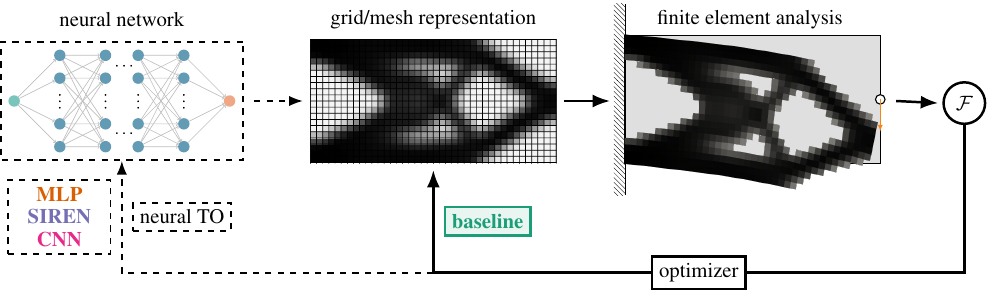}
\caption{Schematic of neural topology optimization (TO). Unlike standard density-based TO (baseline), an NN outputs the physical densities $\boldsymbol{{\rho}}$ (within the bounds $[0, 1]$), on which finite element analysis is performed to obtain the objective. The network parameters are updated through an optimizer to indirectly alter the density field. For the baseline, the individual ``pixels'' are the decision variables, while for the network, trainable parameters form the decision space. Depending on the network architecture, the output can either be the complete density field or the density at a specific location in the design domain. In the latter case, the network represents a continuous field (see Sec. \ref{sec:si_nn} of SI for details about network architectures).}
\label{fig:nrto_workflow}
\end{figure*}

This article focuses on characterizing and explaining the effects of NN reparameterization in topology optimization (TO) \cite{hoyer,tounn,Chandrasekhar2021}, which we name as ``neural topology optimization'' (neural TO). As a baseline for comparison, we use density-based TO with the Solid Isotropic Material with Penalization (SIMP) method \cite{to_first} and the method of moving asymptotes (MMA) as the optimizer~\cite{Svanberg:1987aa}. Note that other TO methodologies could be used as baseline, but the point of this work is not to claim superiority of neural TO over other TO strategies. Instead, we focus on providing the tools to effectively demonstrate both the positive and negative effects of neural TO, guiding the community towards meaningful future developments.

Neural TO uses NNs to nonlinearly transform or reparameterize the design space in the hope of efficiently finding the topology (e.g., material distribution within a computational design domain) that optimizes an objective function (e.g., structural compliance). Fig.~\ref{fig:nrto_workflow} illustrates the method, where it becomes clear that the decision variables $\boldsymbol{\theta}$ are no longer the physical density values $\boldsymbol{\rho}$ at discrete points $\boldsymbol{x}$ (e.g., densities of the finite elements of the discretized design domain). Instead, the decision variables become the NN parameters (weights and biases) and the NN is used to output the physical densities. Then, the corresponding objective is evaluated for that design by a computational solver (e.g., finite element analysis). Optimization constraints (for instance on the maximum allowed amount of material), together with the objective value and its gradient with respect to the decision variables $\boldsymbol{\theta}$ are then provided to an optimizer to improve the design for the next iteration of the optimization. The process repeats until converging to a likely local optimum. Fig.~\ref{fig:nrto_workflow} shows the overall process, where it should be noted that by removing the box titled `neural network', the method reverts to the baseline strategy described earlier (density-based TO). The mathematical formulation of TO and neural TO is detailed in Sec.~\ref{mat_method_sed}.

Neural TO poses a deceptive enigma: Can we improve optimization by reparameterizing the objective landscape with a neural network? At first glance, this strategy is counter-intuitive and the answer might appear to be negative. This is analogous to moving a puppet (objective) by controlling its strings (NN parameters), instead of directly controlling the puppet (density parameters in density-based TO).

\begin{figure*}[!ht]
	\centering    
	\includegraphics{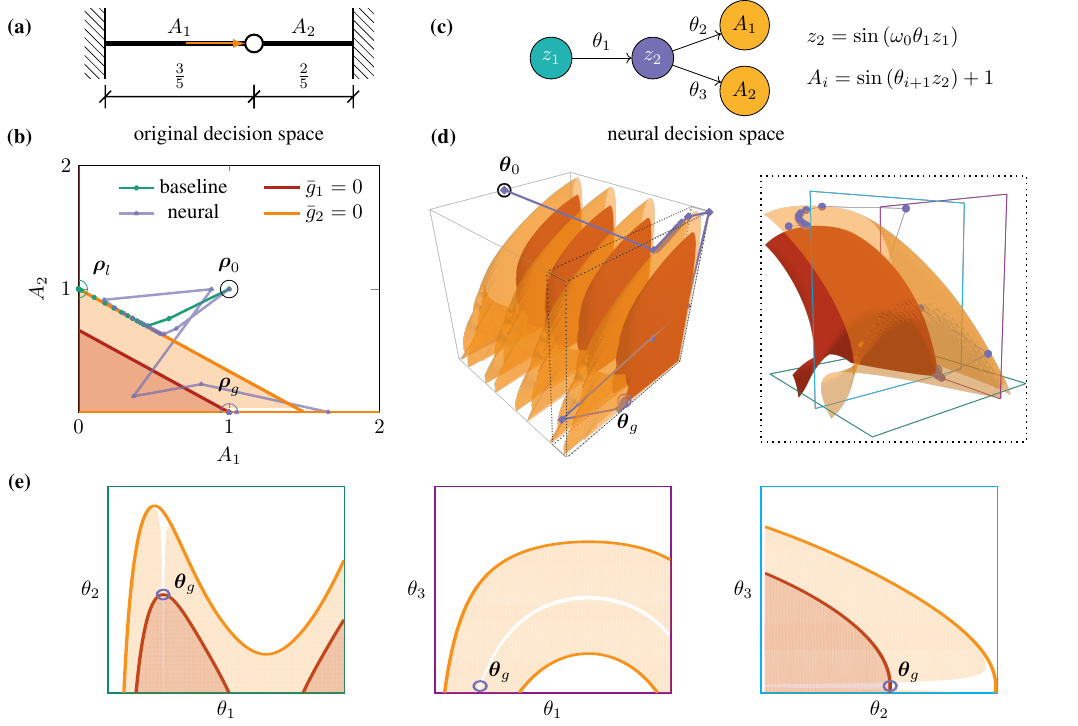}
	\caption{Two-bar problem with stress constraints optimized using MMA, both with and without neural reparameterization: \textbf{(a)} Schematic showing two bars subjected to an axial load at the middle node. The objective is to minimize total mass by varying the bar areas (decision variables $A_1$ and $A_2$), with constraints on each bar's maximum stress; \textbf{(b)} Original decision space, with the white area showing the feasible region of the design space (note the linear feasible subspace near $(1,0)$ along $A_2 = 0$), and the colored regions noting the constraint violations. Starting from a feasible point $\boldsymbol{\rho}_0 = \left( 1, 1\right)$, MMA converges to the local optimum  $\boldsymbol{\rho}_l = \left(0, 1 \right)$. Also shown is the projected trajectory after reparameterization with a SIREN network, converging to the ``singular'' global minimum $\boldsymbol{\rho}_g = \left(1, 0 \right)$; \textbf{(c)} SIREN network used to reparameterize the problem, with a fixed input ($z_1=0.5$) and only 3 parameters ($\theta_i$). The hidden neuron has a parametric $\sin$ activation function, and outputs $A_1$ and $A_2$, where $\omega_0$ is a hyperparameter; \textbf{(d)} The neural decision space and the corresponding trajectory followed by the optimizer in this space. The decision space (left) consists of repeating units, and the inset (right) shows a zoomed view of the distorted constraint surfaces; \textbf{(e)} Three planes passing through the global optimum showing constraints, infeasible regions, and feasible paths to the solution. Details of the plots are given in Sec.~\ref{stress_si} of SI.}
	\label{fig:stress_results}
\end{figure*}

\section{Experiment to illustrate the effect of reparameterization.} Start by considering the stress-constrained optimization problem introduced by Stolpe \cite{stress_to_stolpe}, shown schematically in Fig.~\hyperref[fig:stress_results]{\ref{fig:stress_results}a}. In this simple 2-truss problem with an applied unit load in the middle node, the objective is to minimize the mass of the structure with constraints on the bars' stresses. The decision variables are the areas of the two bars $A_1$ and $A_2$. The mathematical formulation of this problem is included in Sec.~\ref{mat_method_sed}.

This seemingly simple problem has a two-dimensional decision space, which is shown in Fig.~\hyperref[fig:stress_results]{\ref{fig:stress_results}b}. The figure highlights the feasible region (white), the constraints, and local and global optima labeled as $\boldsymbol{\rho}_l$ and $\boldsymbol{\rho}_g$, respectively\footnote{Without constraints, the problem has a trivial global optimum at $(0,0)$. Therefore herein we refer to global optimum that of the constrained problem.}. As apparent from the figure, the design space for this problem is degenerate in the sense that the global minimum $ \boldsymbol{\rho}_g =  \left( 1, 0 \right)$ can be reached from the feasible set only along the line $  A_2 = 0 $. Conversely, the local optimum, located at $ \boldsymbol{\rho}_l =  \left(0, 1 \right)$ can be reached more easily from the feasible set.  It is therefore extremely challenging to find the global minimum using gradient-based optimization. This is shown by the trajectory followed by the MMA optimizer from a  feasible starting point $ \boldsymbol{\rho}_0 = \left( 1, 1 \right)$ (see baseline trajectory (green) in the figure).

To explore the effect of reparameterization, we use the smallest NN architecture consisting of only three weights ($\theta_1$, $\theta_2$, and $\theta_3$) from a four-neuron setup as shown in
Fig.~\hyperref[fig:stress_results]{\ref{fig:stress_results}c}. We remove the network's bias parameters (to facilitate visualization) and nonlinearly transform the original two-dimensional decision space ($A_1$ and $A_2$) into a three-dimensional space. As a result, we show that the same optimizer (MMA) is able to reach the global optimum $\boldsymbol{\rho}_g$ from the same starting point, following the trajectory in Fig.~\hyperref[fig:stress_results]{\ref{fig:stress_results}d}. A two-dimensional projection of this trajectory is also plotted in Fig.~\hyperref[fig:stress_results]{\ref{fig:stress_results}b} for reference, which shows that the optimizer accesses the global optimum through the linear sub-space. Thus, a well-chosen reparameterization reshapes the decision space, making new trajectories possible for the optimizer to reach the otherwise inaccessible global optimum. In this example, two factors facilitate this access. First, in Fig.~\hyperref[fig:stress_results]{\ref{fig:stress_results}d} we see surfaces corresponding to the constraints that are periodically repeated due to the harmonic activation function chosen. As a result, there is an infinite number of global optima. However, this alone does not explain how the optimizer accesses the linear sub-space (which remains degenerate, as we show in Sec. \ref{stress_si} of SI). The inset in Fig.~\hyperref[fig:stress_results]{\ref{fig:stress_results}d} provides a closer view of the path to the optimum, sliced by three orthogonal planes intersecting the found global optimum (Fig.~\hyperref[fig:stress_results]{\ref{fig:stress_results}e}). The feasible regions (white) in these planes indicate that the access path ``opens up''. However, this effect resembles constraint relaxation~\cite{Verbart2016}, where even if the degeneracy persists, it is surrounded by regions with extremely low constraint violations, appearing nearly feasible. Additional analytical and empirical details are provided in Sec.~\ref{stress_si} of SI.

This example illustrates the positive impact of neural reparameterization on nontrivial objective landscapes. However, it does not address the challenges of identifying such beneficial network architectures, nor does it determine whether this approach can be advantageous for more conventional TO problems such as compliance optimization. Additionally, practical neural networks often have orders of magnitude more parameters, complicating the analysis. Therefore, the remaining of the article will discuss a three step analysis strategy to understand the effects of NN choices in the context of neural TO, namely: 1) visualizing objective landscapes; 2) analyzing optimizer trajectories; and 3) quantifying the expressivity of NNs. We select two conventional structural compliance TO problems, namely the tensile and the Michell beam cases (see Fig.~\ref{fig:bc} of the SI), which have optimized solutions with different characteristic features (e.g., coarse and fine features), to showcase the results. We also consider three different NN architectures: 1) a feedforward NN with the commonly used Leaky-ReLU activation function (MLP)~\cite{tounn}; 2) a feedforward NN with sinusoidal activation functions (SIREN)~\cite{Sitzmann2020}; and 3) a convolutional NN (CNN)~\cite{hoyer}. A reader unfamiliar with these NNs is referred to Sec. \ref{sec:si_nn} of SI.

\section{Landscape analysis}
\paragraph{Objective landscape morphology.} 
To compare neural landscapes with the baseline, we adopt the 1-D visualization strategy from ~\cite{goodfellow2014qualitatively}, with modifications to address specific challenges outlined in Sec. \ref{ssec:visualization} of SI. Since all reparameterizations map decision variables to physical densities, we select two physical densities as reference points. We visualize how the landscape varies between these fixed points across different decision spaces.

\begin{figure*}[h]
	\centering
	\includegraphics{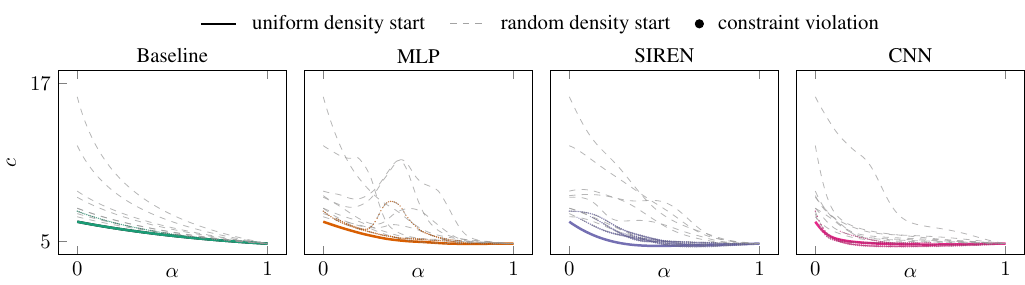}
	\caption{The objective landscapes (interpolating between the same reference points) for different neural reparameterization methods compared against the baseline. The end point, at $\alpha=1$, is the decision space point corresponding to the baseline solution ($\boldsymbol{\hat{\theta}}^\star$) while the starting point is either uniform gray ($\boldsymbol{\hat{\theta}}_u$) or random values (denoted by multiple gray lines). Plots are shown for Michell boundary value problem for SIMP penalty $p=1$ (see Sec. \ref{ssec:visualization} of SI for more results). Constraint violations are indicated by colored markers, with the size of the markers proportional to the violation at each point.}
	\label{fig:NN-MMA-invar}
\end{figure*}

Fig.~\ref{fig:NN-MMA-invar} shows compliance on the ordinate axis as a function of linear interpolation between these two reference points, where $\alpha = 0$ and $\alpha = 1$ denote initial and final designs, respectively\footnote{This linear slice of the landscape reflects what a line-search algorithm would encounter along this direction.}. The first reference point refers to an initial design with uniform density (solid thick line) or random density initial designs (dashed thin lines). The second reference point is the solution obtained from the baseline for $p=1$, known to be a convex problem~\cite{Sigmund2016}. Parameters are then determined to represent these points, and compliance values are evaluated for interpolated parameters $0 < \alpha < 1$. The figure also indicates if the constraint on maximum material has been violated. Fig.~\ref{fig:NN-MMA-invar} clarifies that neural TO leads to non-convex paths that link different initialization points to the final design obtained by density-based TO, i.e., neural TO introduces ``bumps'' in the optimization path when linking the same initial and final designs of the baseline. This is relevant because these ``bumps'' impact the optimization process, as shown in the next subsection. Interestingly, the CNN architecture is less prone to introducing non-convexities than the other two architectures. While the results are shown for $p=1$, the same holds for $p=3$, as well as for all other TO examples we considered (see SI). For $p=3$ some non-convexities are even more pronounced.

\paragraph{Optimizer trajectories.} Visualizing objective landscapes offers qualitative insights into the optimization process, but the linear slices observed do not depict the actual optimizer's path. These landscapes were visualized by connecting initial and final designs (decision variables) obtained through the baseline strategy (not neural TO). Hence, the second step in our analysis contrasts the actual trajectories followed by the optimizer in neural TO versus those of the baseline. This approach uses the optimizer as a probe to explore the landscape. To ensure fairness, we initiate all analyses from the same starting point.

\begin{figure*}[!ht]
	\centering
	\includegraphics{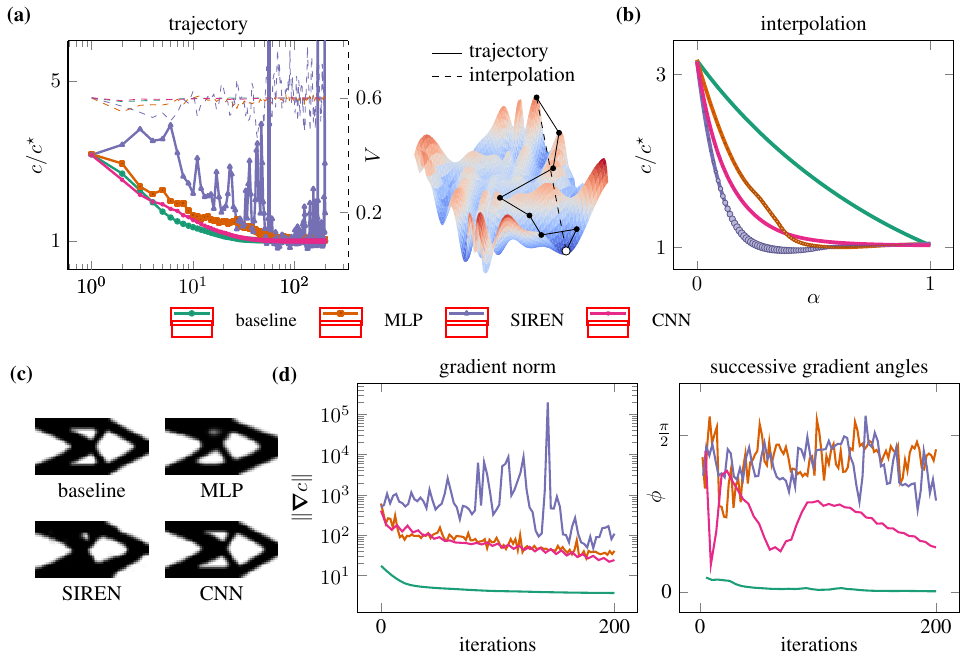}
	\caption{Comparison of MMA's trajectory on the neural landscape against the conventional landscape for the Michell problem (with penalization $p=3$ and target volume fraction of 60\%): \textbf{(a)} Compliance $c$ normalized by the baseline solution $c^\star$ (left ordinate) and volume fraction $V$ (right dashed ordinate), as functions of the optimization iteration; \textbf{(b)} Normalized compliance interpolated between the initial and optimized solutions. The size of the dots indicate the amount of constraint violation; \textbf{(c)} Best feasible designs obtained during optimization for each method, all having similar compliance; \textbf{(d)} $L_2$-norm of the objective gradient and the angle between successive gradient vectors at each point along the optimizer's trajectory.}
	\label{fig:NN-MMA-traj}
\end{figure*}

We consider two different optimizers, such that the distortion of the objective landscape by neural TO is isolated from the choice of optimizer: 1) the method of moving asymptotes (MMA) \cite{Svanberg:1987aa}, which is common in TO literature; and 2) Adam \cite{kingma2014adam}, which is used in the NN literature. Note that hyperparameter optimization is performed for each optimizer fairly for every test case considered (see Sec. \ref{top_opt_details} of SI). Fig.~\hyperref[fig:NN-MMA-traj]{\ref{fig:NN-MMA-traj}a} shows the compliance normalized by the compliance of the best design obtained by the baseline for the Michell problem with $p = 3$, as well as the volume constraint when using MMA with each of the architectures for neural TO. In essence, all cases converge to similar and feasible designs (see Fig.~\hyperref[fig:NN-MMA-traj]{\ref{fig:NN-MMA-traj}c}), although usually requiring more iterations for neural TO compared to the baseline method. Fig.~\ref{fig:NN-MMA-traj} also provides additional information on two important characteristics of the optimization process for neural TO. First, if we linearly interpolate between the initial point in the decision space to the final one, we see in Fig.~\hyperref[fig:NN-MMA-traj]{\ref{fig:NN-MMA-traj}b} that the objective landscape is non-convex for all architectures of neural TO (and for all problems we evaluated, as can be seen in SI). Second, the evaluation of the gradient norm and angle for each iteration in Fig.~\hyperref[fig:NN-MMA-traj]{\ref{fig:NN-MMA-traj}d} reveals that all neural TO strategies zig-zag through the optimization path, i.e., the angle is rarely zero as it was observed for the baseline. This demonstrates that neural TO introduces non-convexities (``bumps'') in the objective landscape (as discussed earlier in Fig.~\ref{fig:NN-MMA-invar}), even for otherwise convex landscapes as obtained for $p=1$. These ``bumps'' perturb the way the optimizer traverses the landscape, thereby slowing down the optimization, i.e., neural TO usually requires more iterations to converge as compared to the baseline method. Fig.~\ref{fig:traj_si} shows the same results when considering the Adam optimizer, which favors the optimization process of neural TO. While using a CNN architecture proved surprisingly competitive, the results herein seem to indicate an overall negative outcome for neural TO since they require more iterations, at least for well-behaved and nearly convex compliance optimization problems. This is an important limitation in practice.

\begin{figure*}[t]
    \centering
    \includegraphics{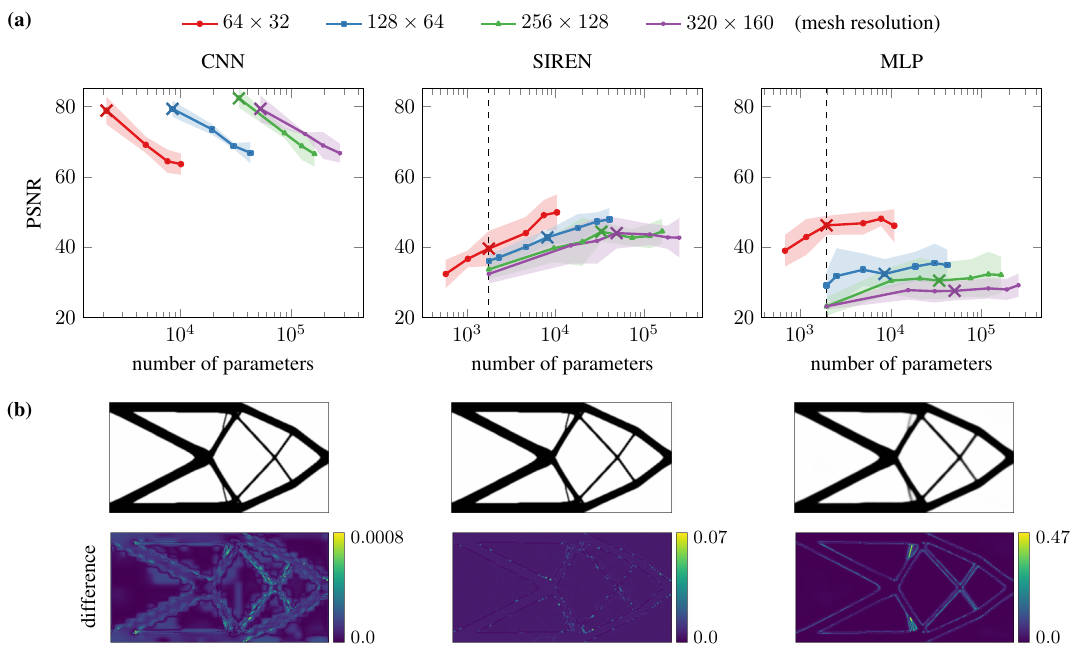}
    \caption{\textbf{(a)} Peak signal-to-noise ratio (PSNR) values for NNs with different number of parameters, measured for 4 mesh resolutions (image sizes). A higher PSNR value indicates that the NN is able to accurately represent the design obtained from the baseline (taken as ground truth). Shaded region denote confidence intervals (one standard deviation) measured across several hyperparameters. The dashed vertical lines correspond to a network that has about $2000$ parameters, independent of mesh resolution. Cross markers indicate architectures with parameters matching each mesh resolution, distinguishing between over-parameterized and under-parameterized regimes. \textbf{(b)} Final designs for the Michell beam problem and their deviations from the baseline for a $320 \times 160$ resolution for all networks corresponding to cross markers.}
    \label{fig:expressivity}
\end{figure*}

\section{Expressivity of neural TO.} The analysis so far focuses on neural network architectures whose number of parameters is comparable to those used by the baseline. However, the design space of neural TO can be over- or under-parameterized, i.e., the number of decision variables (weights and biases of the NN) can be higher or lower than the number of finite elements' density values. Here we investigate the effect of the decision space dimensionality, i.e. number of trainable parameters, for the three NN architectures. If a network is not capable of representing the necessary structural features, then neural TO will not be effective because there will be designs that cannot be created (the design space becomes restricted). Conversely, even if a network is capable of representing the necessary structural features, the objective landscape's non-convexity with respect to the network parameters can make the optimization process more difficult (or easier), as demonstrated earlier.

We assess NN expressivity by assuming the solutions obtained from density-based TO as the ground truth. The network is then trained to learn this image by optimizing its parameters to minimize the pixel-wise error between the ground truth and the network's output (see Sec. \ref{express_study} of SI). Fig.~\ref{fig:expressivity} presents the number of network parameters in abscissas and the peak signal-to-noise ratio (PSNR) in ordinates. PSNR is a common metric for assessing image reconstruction quality, with values above \SI{60}{\decibel} considered high. The PSNR is calculated based on the worst fit obtained from the Messerschmitt-Bölkow-Blohm (MBB) beam, cantilever beam, and Michell beam test cases (see Fig.~\ref{fig:bc}). The CNN has the highest reconstruction capabilities for any number of parameters\footnote{The decreasing PSNR values for increasing CNN parameters is due to two factors: 1) Training with lower floating-point precision on GPUs introduces numerical effects as the errors are already very low ($<10^{-6}$); and 2) Allocating parameters across different types of CNN layers can impact performance.}. Unsurprisingly for a reader familiar with machine learning, MLP and SIREN are less expressive than CNNs. Each curve also includes a cross symbol that separates the under-parameterized regime (to the left) from the over-parameterized regime (to the right). Noteworthy, CNNs are always over-parameterized and both SIREN and MLP lose expressivity as they become under-parameterized. Finally, the expressivity of MLPs drop drastically at higher mesh resolutions, where more fine features are present (Fig.~\hyperref[fig:expressivity]{\ref{fig:expressivity}b}).

The differences in expressivity have been explained in NN literature in the context of image learning \cite{Ulyanov2017DeepIP, nic_inr}. CNNs use convolutional filters that capture important features of images and introduce translation equivariance. While MLPs start training by learning low-frequency features, SIRENs do so with high-frequency features~\cite{rahaman2019spectral}. Practical TO problems often involve slender structures that are more easily reconstructed by networks able to generate high-frequency features.%

 \begin{figure*}[t]
    \centering
    \includegraphics{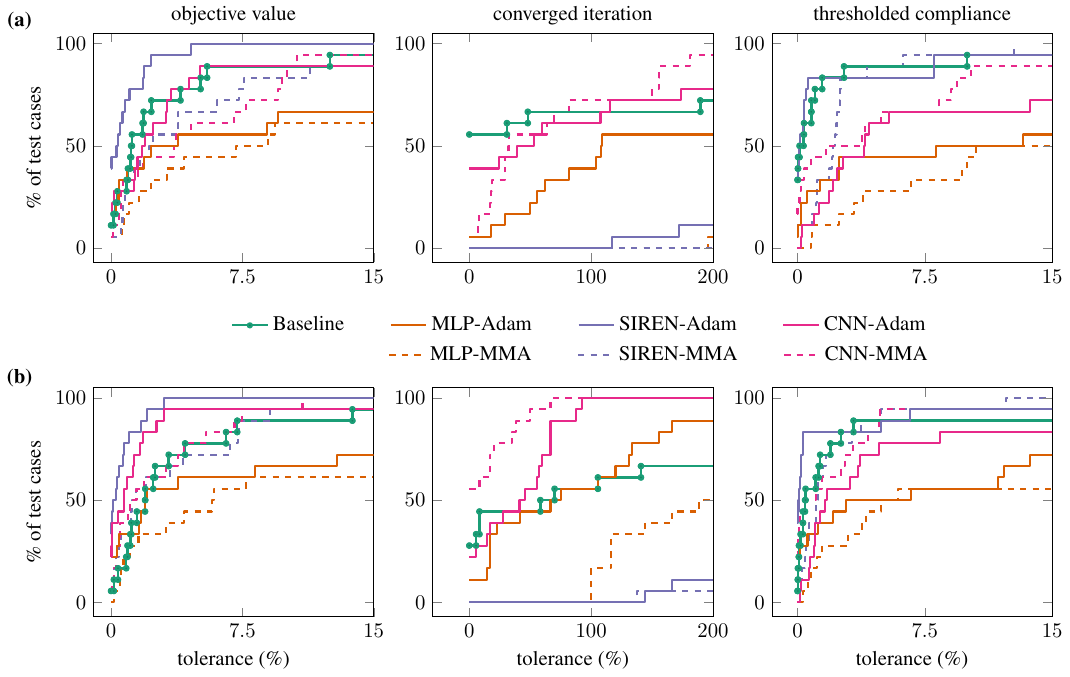}
    \caption{Performance profiles showing \textbf{(a)} Median performance across six different network initializations; and \textbf{(b)} Best performance. For all methods, the profiles compare the best objective value attained during optimization (left column), the number of iterations to convergence (middle column), and the compliance after thresholding to black-and-white designs (right column). The plot shows the percentage of test cases (ordinates) where each method achieved results within a tolerance (abscissas) of the best-performing method for that case. Neural TO was carried out using both Adam and MMA optimizers, while the baseline method only used MMA. All networks have parameters corresponding to the a mesh resolution of $576 \times 288$ elements, and were pretrained to start with a uniform density distribution.}
    \label{fig:overall_pp}
\end{figure*}

\section{Performance of neural TO} 
While a reparameterization's expressivity is beneficial for some boundary value problems, it can be detrimental for others, depending on the problem's features. For example, the density filter used in the baseline reduces expressivity but prevents converging to checkerboard patterns that are otherwise favored by the optimizer. Ultimately, the optimization dynamics determine the effectiveness of a particular reparameterization.

Fig.~\ref{fig:overall_pp} shows the performance profiles for all methods on three optimization problems (namely the MBB, tensile, and bridge cases) and across volume fraction constraints ranging from 10\% to 60\% to ensure feature diversity---totaling 18 distinct problems. The figure evaluates three metrics: the best objective value within a fixed budget of 200 function evaluations (first column), the number of iterations to convergence (second column)\footnote{Defined as the iteration where the objective value saturates, i.e., when the optimizer attains an objective value within a percentage of its corresponding best.}, and the actual performance after thresholding gray designs to black-and-white. The abscissas measure the tolerance percentage, denoting the degree of acceptable deviation from the best solution attained across all methods (not necessarily the baseline), and in ordinates the percentage of problems in which the performance meets this tolerance (see Sec. \ref{top_opt_details} of SI for details).

In the figure, we see that MLPs are less effective, especially for lower volume fractions, because the more intricate optimized designs have slender features that are not favored due to the lack of expressivity. Conversely, CNNs' high expressivity can result in structural links being cut off during thresholding (see the designs for tensile case in Fig.~\ref{fig:overall_designs} and Fig.~\ref{fig:overall_designs_adam}). Additionally, there is concern about performance variability caused by initialization, despite starting from identical initial conditions (e.g., a uniform density).

We also see that Adam finds better solutions than MMA due to the former's ability to navigate the tortuous objective landscapes of neural TO, though this comes at the cost of more function evaluations---particularly for SIREN. Interestingly, both MMA and Adam work well with CNNs, which have a more favorable landscape. When considering the best performance obtained, CNN with MMA has comparable performance to the baseline in terms of objective value and converges faster, contrary to the belief that MMA is unsuitable for optimizing NNs. With the right balance, SIRENs are able to maintain their performance even after thresholding. For more details on performance profiles, see Sec.~\ref{perf_prof} of SI.

\section{Discussion}

``Neural topology optimization'' (neural TO) is based on reshaping or reparameterizing the decision space by including a neural network (NN) before the solver (finite element analysis), and performing optimization on the NN parameters.  We demonstrate that characterizing objective landscapes, optimizer trajectories, and NN expressivity provides a fundamental explanation of the differences in performance between neural TO and density-based TO. Reparameterization changes the way a design is created because the optimizer is acting on new design variables, i.e. the NN's weights and biases instead of directly acting on finite element densities. This follows directly from the fact that a NN has implicit biases, i.e. it has a tendency to create particular types of outputs (designs). For example, MLPs tend to create simpler designs because they favor low-frequency output signals, while SIRENs output designs can favor a wider range of frequencies, thus being capable of also creating slender components. CNNs are particularly balanced compared to MLPs and SIRENs, having better expressivity and being capable of creating very similar designs to the baseline in a small number of epochs (iterations), yet still larger than the baseline. We caution, however, that the final performance of neural TO depends on all three characteristics: the objective landscape that is created by the NN, the way the optimizer traverses it, and the expressive ability of the NN.

However, we note that empirical investigations of performance cannot quantitatively determine if neural TO with a particular NN is \textit{generally better or worse} than the baseline. Performance depends on the problem that is being solved, i.e., it is possible to create sets of problems where each type of reparameterization (neural or otherwise) will outperform a specified baseline due to the ``no free lunch theorem'' in optimization~\cite{Wolpert:1997aa}. Our goal herein was to illustrate why neural TO with particular NN choices can be advantageous or disadvantageous for particular problems, and highlight that using neural TO for originally convex problems has significantly less potential for improvement and may require additional considerations that have not been explored in the literature to date.

At first glance, neural TO may seem a convoluted idea with uncertain benefits. Indeed, we find that reparameterizing the TO problem using a NN can introduce unnecessary complexity (more hyperparameter choices), changes a convex landscape into a non-convex one (or may worsen the landscape of an already non-convex one), and therefore tends to increase the number of iterations required to attain an optimized design for convex problems (compliance optimization). In addition, even if a NN is more expressive than the finite element mesh used to discretize the problem, in the end the output density field from the NN is still evaluated on that mesh, limiting the design capabilities of the NN.
However, we argue that using neural networks has the potential to improve optimization in general, and TO in particular in some circumstances.

First, reparameterization can distort the objective landscape so that it can be traversed differently by the optimizer, thereby enabling the optimizer to reach better optima. We demonstrated this by looking at the two-bar stress-constrained problem. However, while we conjecture that there exists a neural architecture that can be used to reach designs with better performance than the baseline, we caution that finding such architecture is in itself an optimization problem. More research is needed to understand what type of architectures will be broadly suitable to better traverse objective landscapes of most problems. Either way, we postulate that for a wide range of problems reparameterization can be a sound strategy, especially for the ones with nontrivial (non-convex) objective landscapes. In Sec.~\ref{add_results} of the SI, we also show results for thermal conduction and compliant mechanism problems, where neural TO finds better solutions than the baseline.

Second, conventional TO methods do not leverage past knowledge, other than the past experience of the analyst when making modeling choices (optimizer, hyperparameters, etc.). While it is possible to leverage past knowledge using NNs, this is yet not what has been done so far in neural TO literature. Most literature dedicated to using NNs in TO is based on generative machine learning, which attempts to create designs in one shot, i.e., by training a NN on a vast dataset of previously solved TO problems and then hoping that the NN generalizes for unseen problems. In our view, doing this without the ``reparameterization trick'' used by neural TO is not enough because NNs are good interpolatory methods but extrapolate poorly. This can be observed in most articles following the one-shot approach, as this leads to disconnected structures~\cite{ai_review}. Still, we argue that pretraining NNs could offer a significant benefit, e.g., as an effective initialization strategy for neural TO (or even for other methodologies), as they will improve the implicit biases of the NN to adapt and perform well on new problems (extrapolation and generalization). In this article we show that neural TO is already competitive with conventional TO methods using \textit{untrained} NNs, even for convex landscapes. If the NNs have implicit biases that are learned from solving past TO problems, in principle neural TO could outperform conventional TO as long as the past knowledge accumulated in the NN offsets the objective landscape distortions that need to be traversed by the optimizer---an exciting future research direction.

In conclusion, we provided analysis tools for neural TO to explain the difficulties introduced by neural reparameterization in solving otherwise convex problems when using untrained NNs (``the bad''). While we explain the possible slow convergence, we also discuss the potential of using trained NNs to create positive implicit biases.
Furthermore, we highlight that NNs are usually associated with non-convex loss landscapes involving a large number of hyperparameters (``the ugly''). This makes the field empirical, introducing difficulties in terms of replication of results, as well as in fairly comparing the method with conventional TO. We also note that conventional TO has reached giga-voxel resolution~\cite{Aage2017}, while neural TO has not been tested at that scale. While we have considered a maximum resolution of $576 \times 288$ to get insights into neural TO, mega-voxel resolution has been reported elsewhere~\cite{Doosti2021}. However, we also demonstrate the advantages of reparameterization for non-trivial objective landscapes occurring in more complex problems (``the good''). NNs are capable of favorably distorting the objective landscape, which can be useful to reach better optima than those found by conventional methodologies~\cite{Both:2023aa, Heydaribeni2024}. Despite not all NNs being beneficial in the context of TO, they may hold the key to significantly advance the field. Finally, an advantage of carrying out structural optimization on NNs is that we can leverage both software and hardware that have been tailored for machine learning tasks.

\section{Materials and methods}
\label{mat_method_sed}
\subsection*{Density-based topology optimization}
	Formally, the typical TO problem of minimizing structural compliance via a density-based method is formulated as follows~\cite{to_revew}:
	\begin{equation} \label{eq:to_problem}
		\begin{aligned}
			\boldsymbol{\rho}^{\star} = \argmin_{\boldsymbol{\rho} \in \mathbfcal{D} } \quad & \mathcal{F}\left( \boldsymbol{U} \left(\boldsymbol{\rho} \right),  \boldsymbol{\rho} \right) = \boldsymbol{U}^{\intercal} \boldsymbol{F} \equiv c, \\
			\text{such that} \quad & \text{g}_0(\boldsymbol{\rho}) =  V = \textstyle \sum_{i=1}^{N} v_i \rho_i  \leq V_0, \\
			\quad & \boldsymbol{K} \boldsymbol{U} = \boldsymbol{F}, \\
			\quad & 0 \leq \rho_i \leq 1, \quad i = \left\{ 1, \ldots, N \right\}, \\
		\end{aligned}
	\end{equation}
	where $\boldsymbol{\rho} \in \mathbb{R}^N$  is a vector of decision variables (physical density field) taken from the design space $\mathbfcal{D} = [0,1]^N$, $N$ is the total number of finite elements; $\boldsymbol{U}$ is the displacement vector obtained by finite element analysis, i.e., after solving the linear system of equations $\boldsymbol{K} \boldsymbol{U} = \boldsymbol{F}$, with $\boldsymbol{K}$ and $\boldsymbol{F}$ denoting the global stiffness matrix and global force vector, respectively; $c$ is the objective function, which gives a measure of compliance (also related to strain energy), and $g_0$ is the volume of material that needs to satisfy the inequality constraint, with $V_0$ denoting the maximum allowed volume. The subscript $i$ denotes the corresponding element-wise quantities, and thus $v_i$ denotes the volume of the $i$th element and $\rho_i$ its density.  Often, the SIMP law \cite{Bendse2004} is used to penalize intermediate density values (with penalty $p=3.0$) to push the optimization towards a 0 and 1 (black-and-white) design. A density filter is usually applied to prevent the formation of artificially stiff checkerboard patterns (that are often favored by the optimizer) and to enforce a mesh-independent length-scale. The method was re-implemented based on the 88-line Matlab implementation \cite{Andreassen2011}.

\subsection*{Neural topology optimization}
	Introducing a neural network whose output provides the density distribution in a finite element mesh leads to a method we refer herein as ``neural topology optimization'' (recall Fig.~\ref{fig:nrto_workflow}). More broadly, we define \textit{reparameterization} as any approach where the physical density field is expressed as the output of a function $h$, i.e.,  $\boldsymbol{{\rho}} = h(\boldsymbol{\theta})$, where $\boldsymbol{\theta} \in \mathbb{R}^M$ are the new decision variables\footnote{Note that there are many established TO methods that involve a reparameterization (geometric projection~\cite{norato_gp}, moving morphable components~\cite{Guo2014}, Fourier coefficients~\cite{White2018}, wavelets\cite{Poulsen2002}, b-splines\cite{QIAN201315}, and radial basis functions \cite{Wang2005} among others). Even density filters used in density-based TO can be regarded as reparameterizing or transforming the physical densities.}. As a result of reparameterization, the minimized objective and the constraint are  no longer $\mathcal{F}(\boldsymbol{\rho})$ and $g_0(\boldsymbol{\rho})$ but the composition functions $\mathcal{F} \circ h (\boldsymbol{\theta})$ and $g_0 \circ h (\boldsymbol{\theta})$, respectively. Therefore, when an optimizer is used to update $\boldsymbol{\theta}$, the physical density field is altered indirectly. Representing the reparameterization in this way allows for more flexibility in tailoring the decision space, as it can be both over-parameterized ($M > N$) or under-parameterized ($M < N$) when compared to the number of design parameters used in the baseline. Therefore, reparameterization decouples the finite element discretization and the representation of the density field.	
	
	Here we use NNs as reparameterization mappings since they can represent general functions. As a result, the decision space is changed to $\mathbfcal{\widetilde{D}}$, where each parameter becomes unbounded (as opposed to $0 \leq \rho_i \leq 1$ as in the baseline). For the $i$th layer, we carry a mathematical operation of the form $\sigma \left( \boldsymbol{W}_i \boldsymbol{z}_i + \boldsymbol{b}_i \right)$, where $\boldsymbol{z}_i$ is the input, $\boldsymbol{W}_i$ is a matrix of weights, $\boldsymbol{b}_i$ the corresponding bias vector, and $\sigma$ is a nonlinear activation function. For a network of $L$ layers, i.e. $L-2$ hidden layers, the density is calculated as
	\begin{equation} \label{NN_recursion}
		\boldsymbol{\rho} = \sigma \left( \boldsymbol{W}_L \sigma \left( \cdots \sigma \left(\boldsymbol{W}_2 \sigma \left(\boldsymbol{W}_1 \boldsymbol{z}_1 + \boldsymbol{b}_1 \right)
		+ \boldsymbol{b}_2 \right) \cdots \right) + \boldsymbol{b}_L\right).
	\end{equation}
	The reparameterization function $h$ is represented as a NN with trainable parameters $\boldsymbol{\theta}  = \left\{ \boldsymbol{W}_1 \ldots \boldsymbol{W}_L, \boldsymbol{b}_1 \ldots \boldsymbol{b}_L \right\} $, i.e., $\boldsymbol{{\rho}} = h \left(\boldsymbol{\theta}, \boldsymbol{z}_1 \right)$. 

\subsection*{Two-bar stress constrained optimization}
This two-dimensional stress-constrained TO problem is mathematically stated as in~\cite{Verbart2016}:
\begin{equation} \label{two_bar_problem}
	\begin{aligned}
		\boldsymbol{\rho}^{\star} = \argmin_{ \boldsymbol{\rho} \in \mathbb{R}^2} \quad &  \mathcal{F}\left( A_1, A_2 \right) = 0.6A_1 + 0.8A_2, \\
		\text{such that} \quad & \bar{g}_i = \textstyle \left( \frac{A_i}{2} \right)g_i \leq 0, \\
		\quad & 0 \leq A_i \leq 2, \quad i \in \left\{ 1, 2 \right\}, \\
	\end{aligned}
\end{equation}
where the stress constraint is $g_i = \frac{\lvert \sigma_i \rvert}{\sigma_{\max}} - 1$, and $\sigma_{\max} = 1$ is the allowable stress. The objective is to find the values of $A_1$ and $A_2$ that minimize $\mathcal{F}$ according to these constraints. The global optimum is known to be at $A_1=1$ and $A_2=0$, which is a point difficult to reach with the conventional MMA optimizer (in fact, any gradient based optimizer) in this decision space (recall Fig.~\hyperref[fig:stress_results]{\ref{fig:stress_results}b}).

Therefore, we reformulated this problem as a neural TO example considering the smallest NN that reparameterizes the original two-dimensional decision space ($A_1$ and $A_2$) into a three-dimensional decision space ($\theta_1=W_1$, $\theta_2=W_2$ and $\theta_3=W_3$) corresponding to the three weights of the NN shown in Fig.~\hyperref[fig:stress_results]{\ref{fig:stress_results}c}. As shown in that figure, the NN architecture has only 4 neurons: an input neuron $z_1$, a single hidden layer neuron with a sine activation function, and two output neurons that yield $A_1$ and $A_2$. 
We can write this explicitly as:  $A_i = \sin \left( \theta_{i+1} \sin \left( \omega_0 \theta_1 z_1 \right) \right) + 1$, for $i=\left\{1,2 \right\}$, where $\omega_0$ is the frequency hyperparameter. After finding appropriate values for the optimizer's hyperparameters (see Sec.~\ref{stress_si} of SI), the MMA optimizer traverses this 3-D landscape to find the three NN parameters that minimize the objective in Eq.~\eqref{two_bar_problem} subjected to those constraints.

\section*{Acknowledgments}{M.A.B. acknowledges the generous support by the Office of Naval Research through Grant No. N00014-21-1-2670. All authors sincerely thank Dr. Fred van Keulen, Dr. Mathijs Langelaar, Dr. Stijn Koppen, and Dr. Dirk Munro from the Mechanical Engineering faculty at TU Delft for the fruitful discussions and valuable feedback provided. S.M.S. appreciates the efforts of the Bessa research group members for proofreading and pointing out corrections, with special thanks to Gaweł Kus, Igor Kuszczak, and Shunyu Yin for their discussions regarding topology optimization.}

\bibliographystyle{unsrt}  

\newpage

\appendix

\section*{SUPPORTING INFORMATION}

\section{Two-variable stress-constrained problem}
\label{stress_si}
We illustrate the potential benefits of neural reparameterization using the 2-variable stress constrained problem, where we adopt the formulation outlined in \cite{Verbart2016}. The two bars have lengths $L_1=0.6$ and $L_2=0.4$, a unit load $P=1$ is applied at their common node. Notice that the first member is always in tension and the second in compression. We assume a Young's modulus $E=1$, and bar densities $\eta_1=1 $ and $\eta_2=2$.
With this configuration, the stresses in the bars are
\begin{equation*}
    \sigma_1 = \frac{PL_2}{A_1L_2 + A_2L_1}, \qquad \qquad \sigma_2 = -\frac{PL_1}{A_1L_2 + A_2L_1},
\end{equation*}
where $A_i,\ i= \left\{ 1,2 \right\}$ are the bars' areas and also the design (or decision) variables. The objective is to minimize the mass of the structure ($m = 0.6A_1 + 0.8A_2$) subjected to two constraints ($\bar{g}_1, \ \bar{g}_2$) on the maximum allowable stresses ($\sigma_{\max}=1$) of the individual members.

To solve this constrained optimization problem, we used the method of moving asymptotes (MMA)~\cite{Svanberg2002} as the optimizer, with a move limit parameter $m = 2.0$ and the asymptote initialization $a = 0.1$. We tried different values for these parameters (including other starting points) but all of them converged to the local minimum at $(0, 1)$.

\begin{figure*}
    \centering
    \includegraphics{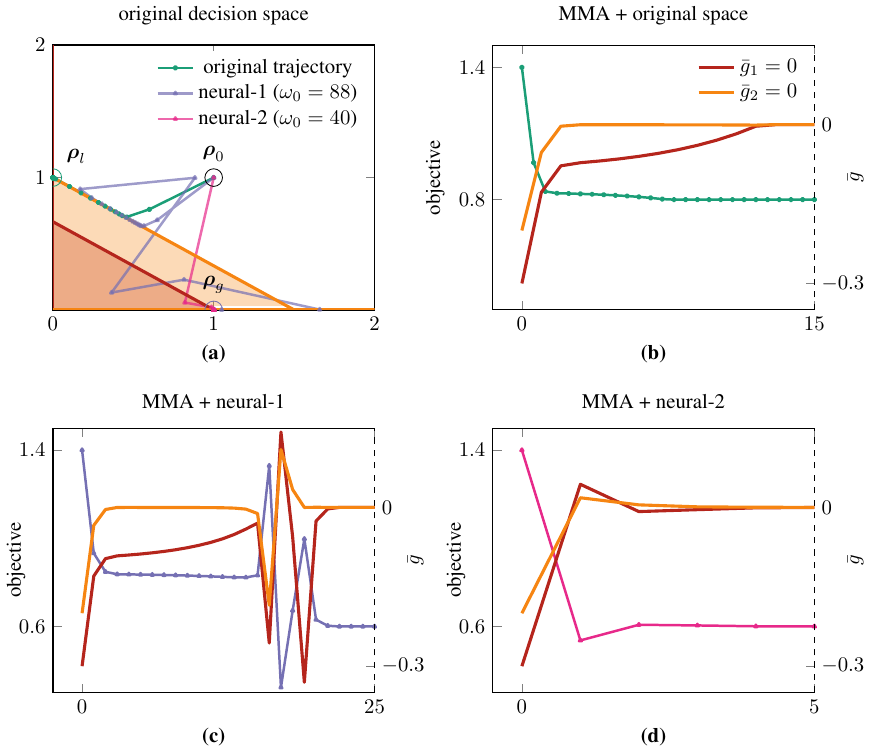}
    \caption{Optimization of the original and reparameterized two-bar stress-constrained problem using a simple SIREN neural network with different frequency parameters $\omega_0$. The MMA optimizer was used for all results; \textbf{(a)} Original 2-D decision space showing the baseline trajectory, and two projected trajectories of the neural optimization. Neural-1 follows the trajectory detailed in the main text ($\omega_0 = 88$), while neural-2 uses a network with $\omega_0 = 40$; \textbf{(b)} Objective (left axis) and constraints (right dashed axis) as a function of optimization iterations for the baseline optimization; and \textbf{(c)-(d)} Similar results for neural-1 and neural-2.}
    \label{fig:stress_si}
\end{figure*}

For the neural reparameterization, we chose a SIREN network architecture with a single input neuron (set to a constant value $z_1 = 0.5$), one hidden neuron with a parametric sinusoidal activation function (i.e., $\sin \left( \omega_0\theta_1 z_1 \right)$), and 2 output neurons yielding the bars' areas $A_i$.
 The latter are obtained through
\begin{equation} \label{eq:reparameterized_areas}
  A_i = \sin \left( \theta_{i+1} \sin \left( \omega_0 \theta_1 z_1 \right) \right) + 1, \qquad i = \left\{ 1,2 \right\},
\end{equation}
so that $0 \leq A_i \leq 2$ (bounds on the original decision variables). Notice that $\boldsymbol{\theta} = \left( \theta_1, \theta_2, \theta_3 \right)$ is the new set of decision variables.
We also use MMA as the optimizer, and its parameters (including the bounds for the variables), together with the frequency parameter $\omega_0$ were tuned using the \emph{Tree-structured Parzen Estimator} algorithm implemented in Optuna~\cite{akiba2019optuna} so as to attain the global optimum for the original problem. For the results shown in the main text we used $m = 0.31$, $a = 0.1$, $\omega_0 = 88$, and the bounds on $\theta$ were set to $\left[ -3, 3 \right]$\footnote{MMA requires bounds on the decision variables for optimization. Since neural weights are unbounded, we have to restrict their range. The choice of bounds is an important hyper-parameter.}. It is worth noting we could find many hyperparameters that converged to the global optimum. Interestingly, for one such set of hyperparameters ($m = 0.4$, $a = 0.3$, $\omega_0 = 40$, and bounds as $[-11, 11]$), the neural reparameterization converged in three iterations (see Fig.~\ref{fig:stress_si}). 

 \begin{figure*}
    \centering
    \includegraphics{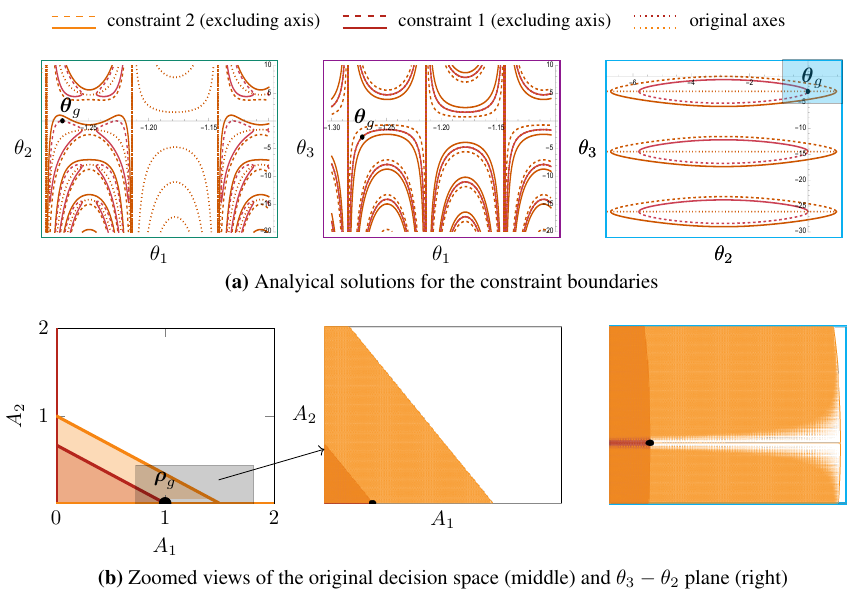}
    \caption{\textbf{(a)} Analytical constraint boundaries in the neural space at three orthogonal planes intersecting the global optimum ($\boldsymbol{\theta}^\star = \left(-1.272, 0, -2.901\right)$) found by MMA optimizer. Dashed and continuous lines of the same color designate the two branches. Dotted lines represent the transformed equations $A_i=0$; \textbf{(b)} illustrates how the original 1-D degenerate subspace to the global minimum is transformed in the neural space. The original decision space is shown on the left (the degenerate subspace has been enlarged for clarity), while the middle figure provides a zoomed-in view of the degenerate subspace (actual scatter plot). The figure on the right shows the same subspace in the neural space. Both plots have the same density of points, and the marker sizes are proportional to the constraint violation.}
    \label{fig:stress_si_analytical}
\end{figure*}

\subsection*{Landscape plots (Fig.~\ref{fig:stress_results} in main text)}
For clarity here we explain how the figure in the main text was constructed. The original constraint boundaries are given by the following equations:
\begin{alignat}{3}
\bar{g}_1 & =0: \qquad && \frac{2}{3}A_1 + A_2 = \frac{2}{3}, \qquad && A_1 = 0, \label{eq:constraint_A1} \\
\bar{g}_2 & =0:  \qquad && \frac{2}{3}A_1 + A_2  = 1, \qquad && A_2 = 0, \label{eq:constraint_A2}
\end{alignat}
where we note that each constraint has two boundaries with $A_i=0$ denoting the coordinate axes.

After reparameterization, the areas are expressed as per Eq.~\eqref{eq:reparameterized_areas}, and therefore Eq.~\eqref{eq:constraint_A1} and Eq.~\eqref{eq:constraint_A2} can be rewritten in terms of $\boldsymbol{\theta}$. We can visually show how these constraint equations are transformed in the neural space, for instance, by looking at their analytical forms for three orthogonal planes intersecting the found global optimum $\boldsymbol{\theta}^\star = (-1.272, 0, -2.901)$ (see Fig.~\hyperref[fig:stress_si_analytical]{\ref{fig:stress_si_analytical}a}).
For the $\theta_3$-$\theta_1$ plane, we can express $\theta_3$ as a function of $\theta_1$ for $\theta_2=0$:
\begin{alignat*}{3}
 & \frac{2}{3}A_1 + A_2 = \frac{2}{3}  \qquad  && \Longrightarrow \qquad \theta_3 = \left(\frac{3\pi}{2} + 2 \pi  c\right) \csc \left(\frac{\omega_0 \theta _1}{2}\right), \qquad && \theta_3 = \left(-\frac{\pi}{2} + 2 \pi  c\right) \csc \left(\frac{\omega_0 \theta _1}{2}\right), \\
& \frac{2}{3}A_1 + A_2  = 1 \qquad  && \Longrightarrow \qquad  \theta_3 = \left(-\arcsin{\frac{2}{3}} + 2 \pi  c\right) \csc \left(\frac{\omega_0 \theta _1}{2}\right),  \qquad && \theta_3 = \left(\pi + \arcsin{\frac{2}{3}} + 2 \pi  c\right) \csc \left(\frac{\omega_0 \theta _1}{2}\right),
\end{alignat*}
\noindent where $\omega_0=88$. It is interesting to note that each constraint now has two branches and there exists an infinite number of constraints for different choices of $c \in \mathbb{Z}$ (the set of all integers). These branches are shown in the figure with dashed and continuous lines, where lines of the same color represent the same constraint boundary. Additionally, the axes, which were also part of the constraint boundaries ($A_i=0$), are now transformed into
\begin{alignat*}{2}
  & A_1 = 0 \qquad && \Longrightarrow \qquad \theta_{i+1} = \left(-\frac{\pi }{2}+2 \pi  c\right) \csc \left(44 \theta _1\right), \\ 
  & A_2 = 0 \qquad  && \Longrightarrow \qquad  \theta_{i+1} = \left(\frac{3 \pi }{2}+2 \pi  c\right) \csc \left(44 \theta _1\right).
\end{alignat*}

and have been represented with dotted lines. We can get equations with similar characteristics for the other planes, i.e., branching into two solutions and being infinitely periodic. However, their expressions are more cluttered, so they are not shown here for brevity but are plotted in Fig.~\hyperref[fig:stress_si_analytical]{\ref{fig:stress_si_analytical}a}. It should be noted that the global optimum must lie on the constraint boundary and is marked on all three planes with $\boldsymbol{\theta}_g$. In Fig.~2e of the main document, these boundaries are plotted over a much smaller range and were identified empirically from data\footnote{A uniform mesh is constructed, and the constraint functions are evaluated at each node. The boundaries are then identified from this data using a marching cubes algorithm.}.

Fig.~\hyperref[fig:stress_si_analytical]{\ref{fig:stress_si_analytical}b} reconstructs parts of the main document figure to explain how the optimizer can access the global constrained optimum in the neural space. On the left, the original decision space is shown schematically, highlighting the 1-D feasible subspace to the global optimum along $A_2=0$. The middle figure presents a scatter plot of a region around the optimum, where marker sizes are proportional to the constraint violation. Values below $10^{-7}$ are treated as feasible (markers disappear), while violations above $10^{-5}$ are considered completely infeasible. Between these values, marker sizes scale logarithmically to the maximum size, resulting in a nearly sharp transition at the meeting point of the two branches of $\bar{g}_2=0$. Here, the thin linear subspace is barely visible. 

For the same plot settings (including the density of scattered points), the $\theta_3 - \theta_2$ plane in the neural space (right) shows a seemingly broader feasible region. However, the analytical plot of the constraint boundaries of the same plane (focusing on the inset at the top-right corner) shows that the only access is along the horizontal dotted line (corresponding to $A_2=0$)\footnote{The reader is reminded that the region inside the ellipses is infeasible, except for this line.}. Thus, while the feasible access is mathematically still one-dimensional, the neural reparameterization induces a ``constraint relaxation'' (similarly to what is discussed in Verbart et al.~\cite{Verbart2016}), creating a surrounding region with negligible constraint violation, thereby facilitating access to the optimum. We made the same plots by lowering the threshold for feasibility (from $10^{-7}$ to $10^{-10}$) and observed no changes.

\section{Neural network architectures} \label{sec:si_nn}

\begin{figure}
  \centering
  \includegraphics{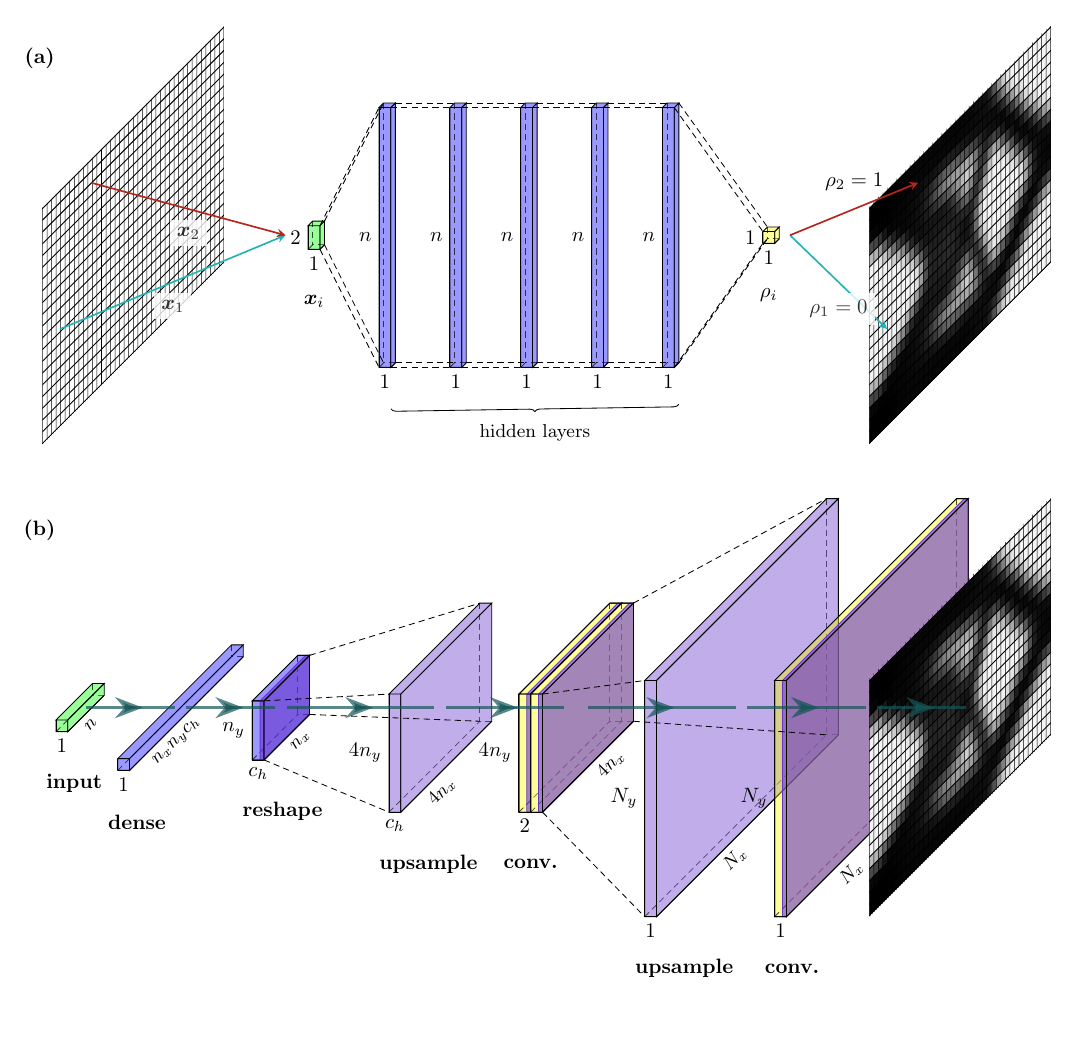}
\caption{Schematic of the neural network architectures considered in this study. \textbf{(a)} shows the fully connected network structure of both MLP and SIREN. The main difference is in their activation functions (Leaky-ReLU and sine respectively). Both MLP and SIREN take Cartesian coordinates ($\boldsymbol{x}_i = \left(x_i, y_i\right)$) and output the physical density at that location ($\rho_i$). As a result, they have to be queried at multiple points in the design domain to form the complete density field ($\boldsymbol{\rho}$). \textbf{(b)} shows the convolutional neural network (CNN) that has a decoder-like structure. It takes an n-dimensional input vector, passes it through a fully connected (dense) layer, and then reshapes it into a small image (proportional to the mesh resolution) with multiple channels ($c_h$). A hyperbolic tangent non-linearity is applied, and the result is scaled up in size using bi-linear interpolation. After normalizing the resulting image by subtracting the mean and dividing by the standard deviation, a convolution operation is applied with two filters, producing two images. These operations (starting from the non-linearity) are repeated once more to obtain the complete physical density field.}
\label{fig:nn_archs}
\end{figure}

The first neural network architecture chosen for this work is the multi-layer perceptron (MLP), where the outputs of neurons in a given layer are connected to all neurons of the next layer (see Fig.~\hyperref[fig:nn_archs]{\ref{fig:nn_archs}a}). Although similar networks have been used by others \cite{Deng2020_tounn2, zehndarntopo, Mai2022, Jeong2023, Qian2022}, we adopt the specific structure chosen by Chandrasekhar and Suresh~\cite{tounn}, which has five hidden layers. The $i$th hidden layer performs a linear transformation $\boldsymbol{l} = W_i\boldsymbol{z}_i + b_i$, batch normalization~\cite{ioffe2015batch}\footnote{A common operation where the output of every neuron is normalized by rescaling and recentering based on all inputs of the network.}, and applies the Leaky-ReLU nonlinear activation $\sigma = \max{(0.01\boldsymbol{l},\ \boldsymbol{l})}$, where the function is applied element-wise.

A limitation of MLP networks is their difficulty in representing high frequency functions, which is known as \emph{spectral bias}~\cite{rahaman2019spectral}. This issue can be addressed by approaches that transform inputs using random Fourier features~\cite{tancik2020fourier},or by using the sine function as activation function. These networks are known as SIRENs~\cite{Sitzmann2020}. Although Chandrasekhar and Suresh~\cite{Chandrasekhar2021}, and Doosti et al.~\cite{Doosti2021} used the random Fourier features approach alongside a single hidden layer MLP, we choose SIRENs for this work due to their structural and parametric similarity to MLPs. A hidden layer of a SIREN takes the input and applies the transformation $\sin \left( \omega_0 \left( W_i\boldsymbol{z}_i + b_i \right) \right)$, where $\omega_0$ is a hyperparameter that dictates the highest frequencies that can be represented.

SIRENs and MLPs are collectively known as \textit{coordinate-based networks} or \textit{implicit neural representations}.
The coordinates of the centers of the finite elements $\boldsymbol{x}_i$ are given as inputs to the network and their corresponding density values $\rho \left( \boldsymbol{x}_i \right)$ is retrieved from the network. For a 2-D finite element mesh, this mapping can be written as $\rho_i = h \left( \boldsymbol{\theta}; \boldsymbol{x}_i \right): \mathbb{R}^2 \rightarrow \mathbb{R}$. Noteworthy, the network outputs are constrained using either a sigmoid or a shifted-sigmoid transformation to bound the physical densities so that $0 \leq \rho_i \leq 1$. 

Finally, we also consider a convolutional NN (CNN), an architecture widely used for image processing. Specifically, for this study we use the decoder-type architecture adopted by Hoyer et al.~\cite{hoyer} and by Zhang et al.~\cite{Zhang2021_tonr}, which takes as input a vector $\boldsymbol{z} \in \mathbb{R}^n$ to generate the physical density field $\boldsymbol{\rho}$. The input vector---whose dimension $n$ is an architectural choice, often chosen such that it is less than the number of image pixels---is treated as a trainable parameter and is randomly initialized; the CNN's mapping $\boldsymbol{\rho} = h \left( \boldsymbol{\theta}; \boldsymbol{z} \right): \mathbb{R}^{n} \rightarrow \mathbb{R}^{N}$, yields the full density field of the finite element discretization. Here, $N = N_x \times N_y$, with $N_x$ and $N_y$ denoting the number of finite elements along their respective Cartesian directions.
Unlike MLPs and SIRENs, the CNN's architecture (see Fig.~\hyperref[fig:nn_archs]{\ref{fig:nn_archs}b}) depends on the mesh resolution, i.e., once the architecture is chosen, the resulting design is fixed in size. Therefore, to generate a larger resolution design, the number of CNN parameters has to be scaled up (similarly to the baseline). After transforming linearly the input and applying a nonlinear activation function (specifically the hyperbolic tangent), the resulting vector is reshaped to a 3-D tensor of size $n_x \times n_y \times n_c$, where $n_x$ ($n_y$) is proportional to $N_x$ ($N_y$) but smaller by a factor of $32$, and $n_c$ is the number of channels required (another architectural choice). Consequently, this tensor can be viewed as a small image with multiple channels, allowing for the following image-based operations:
\begin{itemize}
    \item \textit{Bi-linear upsampling} performs neighborhood interpolation to upscale an image without any parameters;
    \item \textit{Normalizing} centers an image's pixels by subtracting the mean value of the image and dividing by the standard deviation;
    \item \textit{Convolution} performs a weighted average of the neighborhood of each pixel using trainable filters. Each filter has the same number of channels as the image it is acting on and it is chosen to have a $3 \times 3$ size, with each filter producing one output image;
    \item \textit{Offsetting} is an operation where trainable parameters, with the same spatial shape as the image, are added to the image; effectively this operation acts as bias to shift each pixel's intensity or value.
\end{itemize}

Each hidden layer consists of these operations, followed by a $\tanh$ activation function. As opposed to the architecture used in \cite{hoyer, Zhang2021_tonr} that had 5 hidden layers, we limit the number of hidden layers to two. Additionally, in experiments designed to match the CNN architecture’s number of parameters with the baseline, we set $n_c=1$, $n=1$, and used only three convolution filters: two in the first hidden layer and one in the last.

Other choices were made to make fair comparisons: We do not use continuation schemes for the SIMP penalty factor. Furthermore, although NNs become more expressive either through increasing the number of hidden layers (depth)~\cite{lu2017expressive} or the number of neurons per layer (width)~\cite{UAT}, we chose the latter and fixed the depth for all experiments. All networks were implemented in JAX \cite{jax2018github} using the Haiku library~\cite{haiku2020github}, within our in-house topology optimization library.

\subsection{Volume constraint enforcement} \label{app:hoyer_sigmoid}
Most publications that use reparameterization (e.g., \cite{tounn, Chandrasekhar2021} among others)  either use quadratic penalty or augmented Lagrangian methods to convert the constrained optimization problem to an unconstrained one, thereby enabling common optimizers in machine learning to be used. However, these introduce additional hyperparameters, such as the initial penalty value, its increment magnitude per iteration, and the criterion used to increase the penalty value. Setting these hyperparameters appropriately can be challenging. Instead, we either use MMA as the optimizer for the neural TO---for which enforcing the volume constraint is straightforward---or when using Adam as the optimizer we enforce the volume constraint through the shifted-sigmoid strategy~\cite{hoyer}. The latter applies a parametric-sigmoidal transformation on the outputs of the network at each iteration of the optimization process. Briefly, we use the following sigmoid function:
\begin{equation*}
    \rho_i = \frac{1}{1 + \exp^{\left(\rho_i - b\left(\boldsymbol{\tilde{\rho}}, V_0\right)\right)}},
\end{equation*}
\noindent where the output of the network (i.e., $\boldsymbol{\tilde{\rho}}$) is transformed into physical densities bounded between 0 and 1. The scalar parameter $b$ shifts the output to ensure the volume constraint is enforced exactly at every iteration, i.e., $g_0(\boldsymbol{\rho}) = V_0$. This converts the original constrained problem into an unconstrained one, allowing the use of common ML optimizers. The value of $b$ depends on the output and the required volume fraction, and it is determined using a bisection algorithm (within an error of $10^{-12}$). For the neural networks, we obtained smoother results when this operation was carried out after the density filtering. 

\section{Landscape analysis}
\label{sec:si_landscape}

\begin{algorithm}[t]
  \caption{Loss landscape visualization using linear interpolation}
  \label{alg:loss_vis}
  \begin{algorithmic}[1]
    \Require Reparameterization maps: $\mathcal{H} = \{h_1, \dots, h_n\}$, reference points in the physical density space $ \boldsymbol{{\rho}}_1, \boldsymbol{{\rho}}_2$
    \Ensure Sets $\mathbfcal{F}$ (function values) and $\mathbfcal{G}$ (constraint values)
    
    \Statex
    
    \State \inlinecomment{Obtain decision space points}
    \For{$h_i \in \mathcal{H}$}
    
      \Let{$\boldsymbol{\hat{\theta}}_1$}{$\argmin_{\boldsymbol{\theta}} \frac{1}{N} \norm{ h_i \left( \boldsymbol{\theta} \right) - \boldsymbol{{\rho}}_1 }^2 $} \Comment{Obtain first point in decision space}

      \Let{$\boldsymbol{\hat{\theta}}_2$}{$\argmin_{\boldsymbol{\theta}} \frac{1}{N} \norm{ h_i \left( \boldsymbol{\theta} \right) - \boldsymbol{{\rho}}_2 }^2 $} \Comment{Obtain second point in decision space}

      \State \inlinecomment{Interpolate between points}

      \For{ $0 < \alpha < 1$ }
        \Let{$\boldsymbol{\theta}_\alpha$}{$\boldsymbol{\hat{\theta}}_1 + \alpha \left( \boldsymbol{\hat{\theta}}_2 - \boldsymbol{\hat{\theta}}_1 \right)$} \Comment{Obtain intermediate parameter point}
 
        \Let{$\mathbfcal{F}$}{$\mathbfcal{F} \cup \left( \alpha,  \mathcal{F} \circ h_i \left( \boldsymbol{\hat{\theta}}_\alpha \right) \right) $} \Comment{Evaluate function value and store pair in set}
        \Let{$\mathbfcal{G}$}{$\mathbfcal{G} \cup \left( \alpha,  g_0 \circ h_i \left( \boldsymbol{\hat{\theta}}_\alpha \right) \right)$} \Comment{Evaluate constraint value and store pair in set}
         
      \EndFor
    \EndFor
    \State \Return {$\left\{ \mathbfcal{F}, \mathbfcal{G} \right\} $}
    
  \end{algorithmic}
\end{algorithm}

\subsection{Visualization}
\label{ssec:visualization}

High dimensional neural loss landscapes are often visualized by either taking 1-D slices~\cite{goodfellow2014qualitatively} or by projecting into 2-D planes~\cite{li2018visualizing}. While the latter has been used to explain the effects of certain architectural choices, the high-dimensional space is visualized along two random directions \footnote{By drawing random numbers from a Gaussian distribution, the chosen vectors are nearly orthogonal in high-dimensional space. This is a reasonable assumption since networks visualized using this approach often have millions of parameters.} by perturbing a single point. This makes it difficult to compare the baseline with the neural schemes: First, unlike neural network parameters, the decision variables in the baseline are bounded, and thus random perturbations can violate the bounds, making the loss evaluation impossible, especially near minima. Second, the resulting visual representations are difficult to interpret since the plot is made along random directions. Finally, and most importantly, the magnitude of the perturbations is arbitrary. As a result, smaller perturbations often show convex landscapes (due to the local convexity in even non-convex functions). Thus, comparing loss landscapes without considering the length-scale of the visualization may be meaningless.

We propose a simple modification to make 1-D visualizations comparable across different landscapes by relying on the general definition of reparameterization as mappings from the decision space to the physical density space (i.e., $\boldsymbol{\rho} = h(\boldsymbol{\theta})$). The procedure is detailed in Algorithm~\ref{alg:loss_vis}.
First, we choose two reference points $\boldsymbol{\rho_1}$ and $ \boldsymbol{\rho_2}$ in the physical density space. For this study, we chose the physical density of the baseline's converged solution as the first reference point. For the second point, we investigated two options:
\begin{enumerate}
  \item A uniform density design, which corresponds to the commonly used starting point of the baseline; and
  \item Multiple random density designs, which are commonly used as initialization points for the neural networks.
\end{enumerate}
Second, for each reparameterization, we find the points in the decision space that generate these densities by solving the following optimization problem:
\begin{equation}
   \boldsymbol{\hat{\theta}} = \argmin_{\boldsymbol{\theta}} \:  \frac{1}{N}\sum_{j=1}^N \left( h(\boldsymbol{\theta})_j -{\rho}_j \right)^2,
\label{eqn:regression_opt}
\end{equation}
\noindent where $N$ is the number of finite elements\footnote{Note that even though we have used $\boldsymbol{\theta}$ here, this method is equally applicable to visualize the baseline's landscape as well. For instance, it can be used to assess the effect of adding projection filters or making changes to SIMP's penalty, among other modifications. Furthermore, other reference points can also  be chosen based on convenience.}. By solving Eq.~\eqref{eqn:regression_opt} once for each reference point and for a given reparameterization $h$, we obtain the corresponding decision space points $\boldsymbol{\hat{\theta}_1}$ and $ \boldsymbol{\hat{\theta}_2}$. Visualization in 1-D works by interpolating between these two points, i.e., by evaluating the objective and constraint values at a series of points between them. If the two points chosen have decision variables that are within bounds, all points along the line joining them would also satisfy the bounds. Mathematically, any point on the line joining two points $\boldsymbol{\theta}_1$ and $\boldsymbol{\theta}_2$ can be represented as
\begin{equation} \label{eq:linint}
    \boldsymbol{\theta}_\alpha = \boldsymbol{\theta}_1 + \alpha \left( \boldsymbol{\theta}_2 - \boldsymbol{\theta}_1 \right), \quad 0 \leq \alpha \leq 1.
\end{equation}
Therefore, the loss landscape can be plotted by calculating the objective $\mathcal{F}\circ h(\boldsymbol{\theta}_\alpha)$ and constraint $g_0 \circ h \left( \boldsymbol{\theta}_\alpha \right)$ values for different values of $\alpha$. By keeping the density space points the same for different reparameterizations, the loss landscapes' length-scales are linked and comparisons are fair.

\begin{figure}
\resizebox{\textwidth}{!}{
 \includegraphics{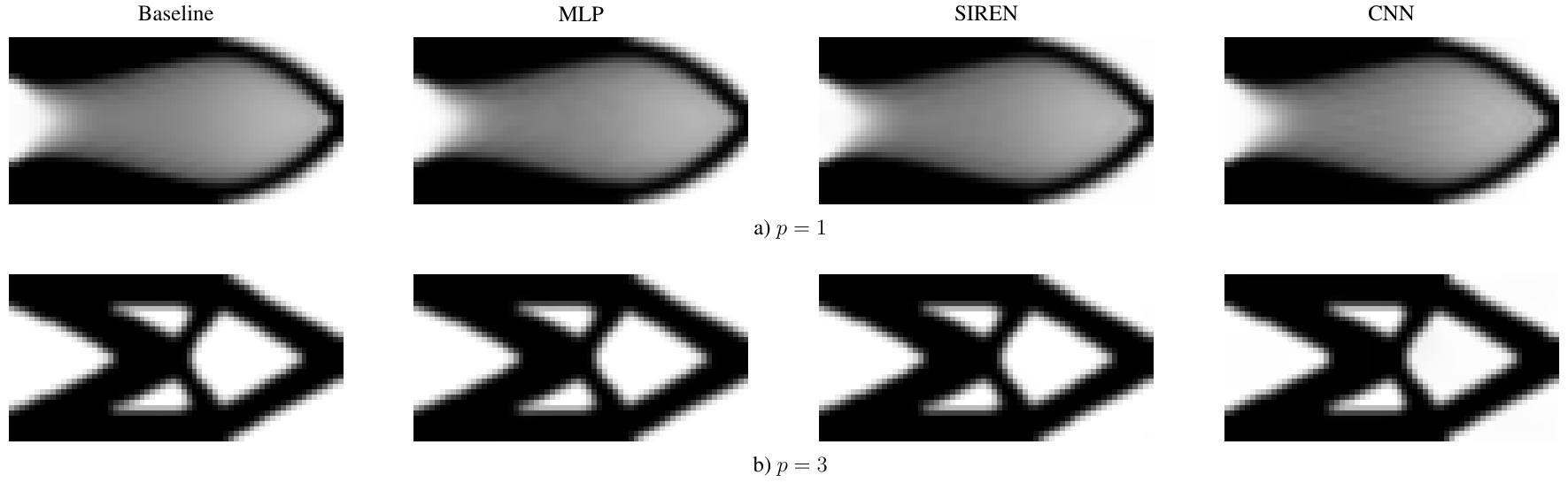}
 }
 \caption{The physical densities obtained for the Michell test case after solving Equation \ref{eqn:regression_opt} for the reference point (corresponding to the solution from the baseline).}
 \label{fig:app_reg_designs}
\end{figure}

\begin{figure*}
  \centering
  \includegraphics{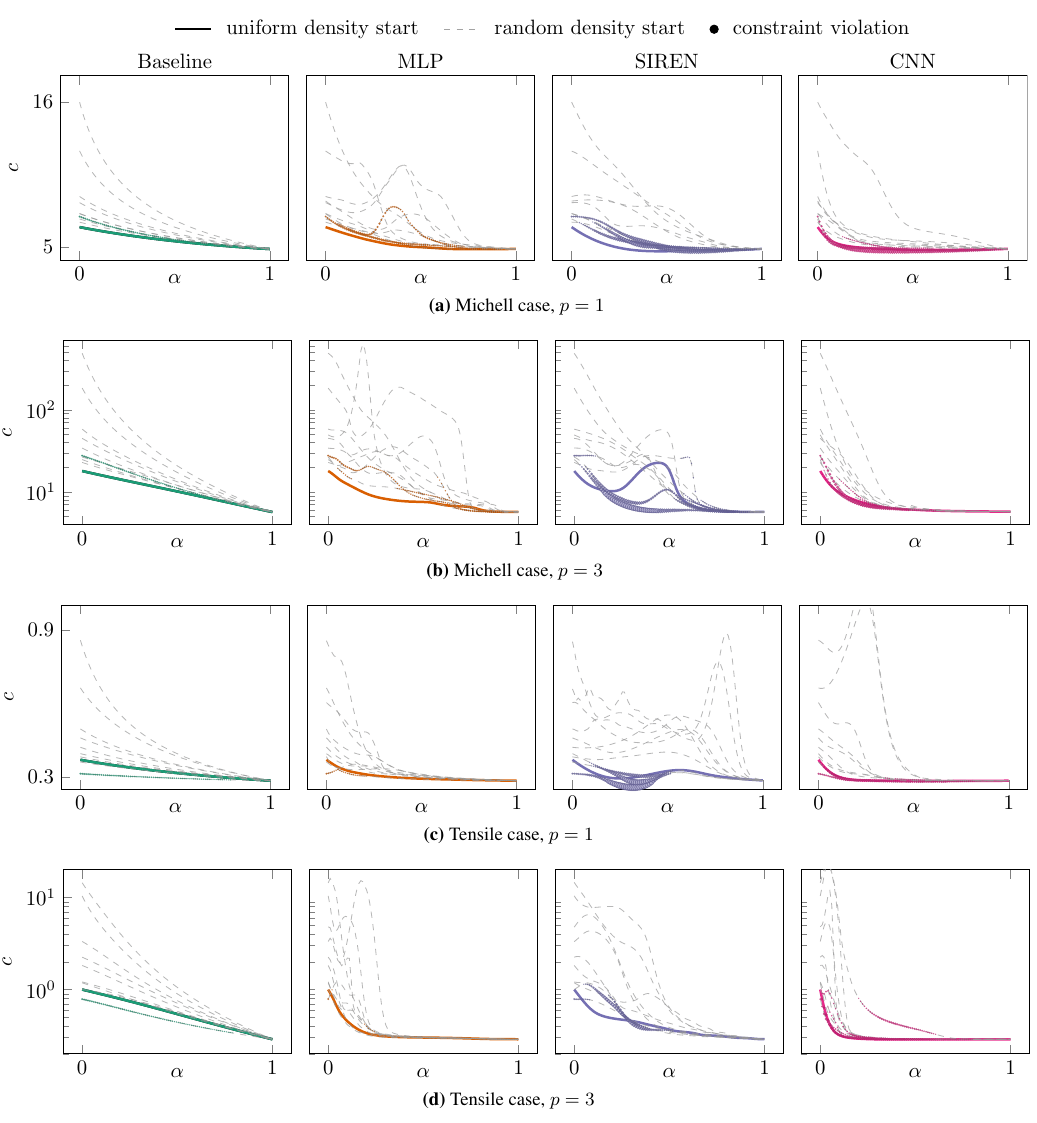}
   \caption{Loss landscapes (interpolating between the same reference points) for different neural reparameterization methods compared against the baseline. The end point, at $\alpha=1$, is the decision space point corresponding to the baseline solution while the starting point is either uniform gray or random values (denoted by multiple gray lines). The first row has been shown in the main text.}
   \label{fig:NN-MMA-invar_si}
\end{figure*}

To remove the bias introduced by the dimensionality of the design space, we chose network architectures (for MLP, SIREN, and CNN) with roughly the same number of parameters as the baseline. Thus, MLP and SIREN have 20 and 22 neurons per layer in 5 hidden layers, totaling \num{1961} and \num{2113} parameters, respectively; these are  within 5\% of the baseline's parameter count (\num{2048}). As for the CNN, the original architecture was simplified by reducing the number of hidden layers to two and by having only one trainable input ($n=1$), a single channel ($n_c=1$), and 2 convolution filters, yielding \num{2156} parameters. The L-BFGS optimizer implemented in JAXOPT~\cite{jaxopt_implicit_diff}, with default hyperparameters, was used to solve the optimization problems. Table~\ref{tab:visual_errors} gives the maximum errors obtained during this process for each of the reparameterizations. In addition, for reference, the physical density fields corresponding to the solutions of the optimization problem~\ref{eqn:regression_opt} are shown in Fig.~\ref{fig:app_reg_designs}. It can be seen that the designs are virtually indistinguishable from one another. Fig. \ref{fig:NN-MMA-invar_si} shows the results of visualizing between the initialization point at $\alpha=0$ (either uniform or random) and the baseline solution at $\alpha=1$. Note that this method can be used to see the effect of changes to landscape induced by the addition of filters, the choice of different activation function, the number of layers, among others.

\begin{table}[h!]
\centering
\caption{Maximum mean square error during optimization (Equation \ref{eqn:regression_opt}) for loss landscape visualization.}
\begin{tabular}{@{}lllll@{}}
\toprule
                       & \textbf{Baseline} & \textbf{MLP} & \textbf{SIREN} & \textbf{CNN} \\ \midrule
Tensile ($p=1$) & $6.6\times 10^{-8}$            & $5.6\times 10^{-5}$       & $1.4\times 10^{-5}$         & $3.9\times 10^{-7}$       \\
Michell ($p=1$) & $9.6\times 10^{-8}$            & $7.3\times 10^{-5}$       & $3.3\times 10^{-5}$         & $2.5\times 10^{-5}$       \\
Tensile ($p=3$) & $7.2\times 10^{-8}$            & $5.6\times 10^{-5}$       & $3.4\times 10^{-5}$         & $3.9\times 10^{-7}$       \\
Michell ($p=3)$ & $1.6\times 10^{-7}$            & $5.6\times 10^{-5}$       & $1\times 10^{-5}$           & $4.2\times 10^{-4}$       \\ \bottomrule
\end{tabular}

\label{tab:visual_errors}
\end{table}

\begin{figure*}
  \centering 
  \resizebox{\textwidth}{!}{
  \includegraphics{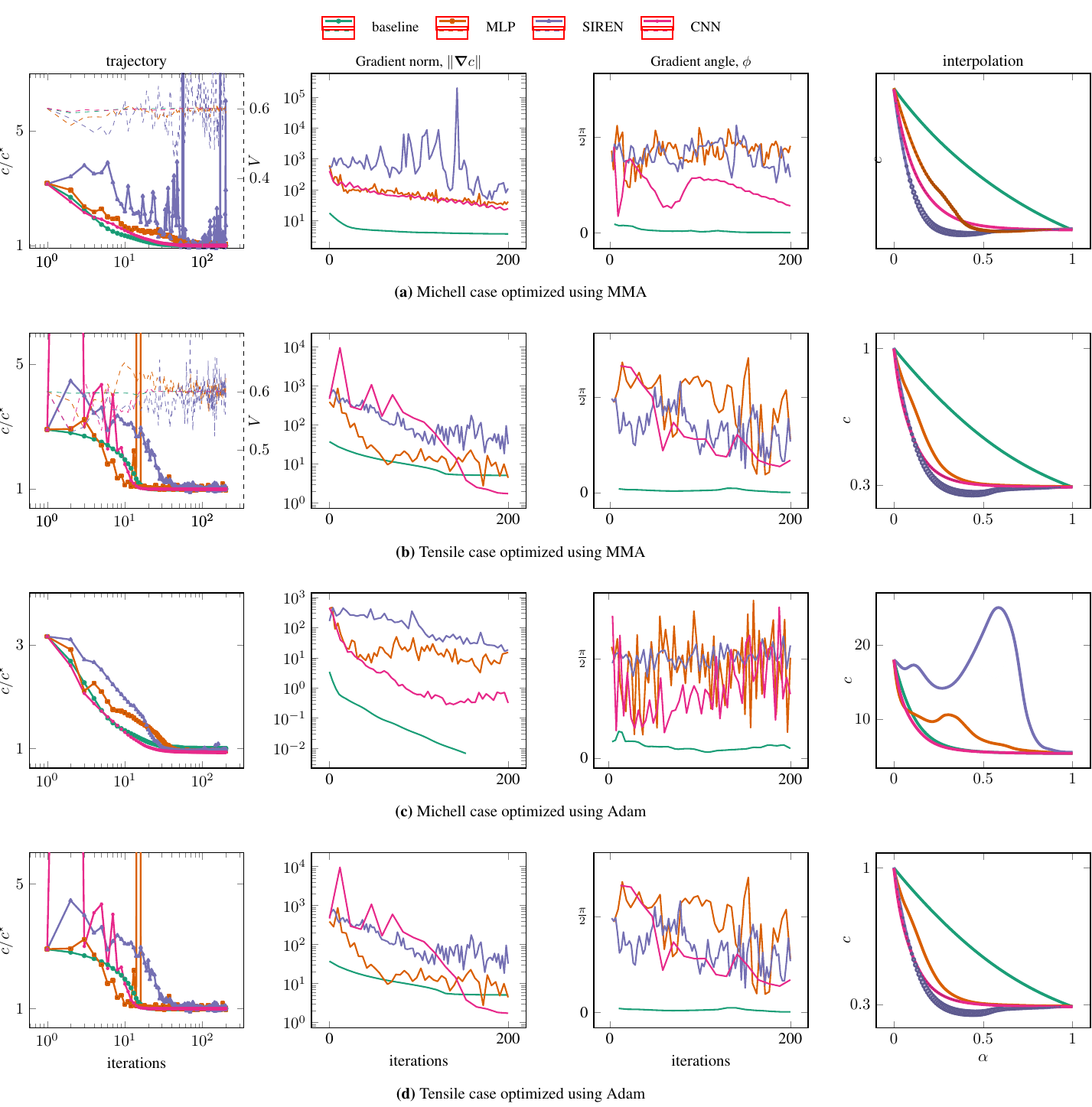}
  }
   \caption{Comparison of optimizer's trajectory on the neural landscape against the conventional landscape (with penalization $p=3$ and target volume fraction of 60\%). Left column shows compliance $c$ normalized by the baseline solution $c^\star$ and volume fraction $V$, as functions of the optimization iteration; The second column is the normalized compliance along interpolation between initial and final designs. The size of the dots indicate the amount of constraint violation; $L_2$-norm of the objective gradient and the angle between successive gradient vectors at each point along the optimizer's trajectory are shown in the last two figures.}
   \label{fig:traj_si}
\end{figure*}

\subsection{Trajectory analysis}\label{traj_analysis}
By using the same optimizer, the differences in the trajectories can reveal how the reparameterization alone affects the landscapes. To compare the trajectories, we note that the gradient of the objective function $\grad c$ plays an important role in how the optimizer traverses the landscape. Therefore, we plot the magnitude of the objective's gradient $\norm{\grad c}$---calculated using automatic differentiation and verified using finite difference calculations---and the gradient's direction. The latter is calculated as the angle $\phi$ between successive gradient vectors according to the cosine similarity, i.e.,
\begin{equation*}
    \cos \phi = \frac{\grad c^{(i)} . \grad c^{(i-1)}}{\norm{\grad c^{(i)}}\norm{\grad c^{(i-1)}}},
\end{equation*}
\noindent for iterations $i-1$ and $i$. 
In addition, we plot the optimization history, showing how the objective and constraints vary from one step of the optimizer to the next in the different landscapes. Finally, we examine the linear subspace that connects the starting and ending points of the optimization trajectory. To do this, we use Eq.~\eqref{eq:linint} to interpolate between the starting point (uniform gray, which remains the same for all methods) and the solutions found in each decision space. It is important to note that such a plot does not allow comparisons among landscapes but provides information specific to the converged solution, as Eq.~\eqref{eqn:regression_opt} is not used for making these plots.

Fig.~\ref{fig:traj_si} shows these metrics for all models, test problems (tensile and Michell), and optimizers (MMA and Adam). When considering the actual trajectory of the optimizer (first column of the figure), the baseline has a smooth objective curve, while the network's trajectories are noisy, i.e., even small steps may result in large increases of the objective function value. Furthermore, for the networks, MMA struggles to satisfy the constraint. This is expected since the linear constraint has now been transformed into a non-convex one ($g_0 \circ h$) due to the presence of the nonlinear reparameterization. The gradient information (second and third columns) is also very noisy: The norm can vary widely from step to step, indicating a chaotic search. Additionally, the angle between successive gradients shows that the trajectory is almost perpendicular (or even in opposite directions) from one iteration to the next, especially for SIREN and MLP architectures. This phenomenon is independent of the optimizer used, suggesting that it is an inherent effect of the network landscape. In contrast, the baseline's gradients point consistently in one direction, allowing smooth convergence to the minimum. These results suggest that the complex neural landscapes may hinder fast convergence, as the local information that the optimizer uses to make steps is not sufficiently informative. Moreover, the objective landscapes of neural TO may not be effectively traversed by optimizers used in conventional TO (e.g., MMA).  Neural network optimizers like stochastic gradient descent (SGD) or Adam may be more appropriate. Finally, the linear interpolation (second column) is a smooth, feasible, monotonically decreasing curve for the baseline. However, for the neural networks, the linear subspace to the converged solution has locations of either high objective values or large constraint violations.

\begin{table}[h!]
\centering
\caption{Details of the neural architectures used to study expressivity. For fully connected networks with five hidden layers, the width refers to the number of neurons per layer. For CNNs, ``Input size'' refers to the dimensionality of the trainable input, ``Channels'' to the number of initial dense channels, and ``Filters'' to the number of convolution filters in the first layer. The CNN has two hidden layers. Note that the first two architectures are under-parameterized, making them not applicable for CNN.}
\label{tab:app_exp_table}
\begin{tabular}{cccccc}
\hline
\multirow{2}{*}{\textbf{Mesh resolution}} & \multirow{2}{*}{\textbf{SL No:}} & \multicolumn{3}{c}{\textbf{CNN}}                           & \textbf{SIREN and MLP} \\ \cline{3-6} 
                                          &                                  & \textbf{Input size} & \textbf{Channels} & \textbf{Filters} & \textbf{Width}         \\ \hline
\multirow{6}{*}{$64 \times 32$}   & 0 & \xmark  & \xmark & \xmark  & 11  \\
                         & 1 & \xmark  & \xmark & \xmark  & 15  \\
                         & 2 & 1   & 1  & 2   & 20  \\
                         & 3 & 16  & 12 & 16  & 33  \\
                         & 4 & 32  & 12 & 32  & 42  \\
                         & 5 & 64  & 16 & 32  & 50  \\ \hline
\multirow{6}{*}{$128 \times 64$}  & 0 & \xmark  & \xmark & \xmark  & 23  \\
                         & 1 & \xmark  & \xmark & \xmark  & 33  \\
                         & 2 & 1   & 1  & 2   & 44  \\
                         & 3 & 32  & 12 & 32  & 66  \\
                         & 4 & 64  & 12 & 64  & 85  \\
                         & 5 & 128 & 16 & 64  & 100 \\ \hline
\multirow{6}{*}{$256 \times 128$} & 0 & \xmark  & \xmark & \xmark  & 48  \\
                         & 1 & \xmark  & \xmark & \xmark  & 70  \\
                         & 2 & 1   & 1  & 2   & 90  \\
                         & 3 & 64  & 16 & 32  & 135 \\
                         & 4 & 96  & 16 & 64  & 170 \\
                         & 5 & 128 & 16 & 96  & 200 \\ \hline
\multirow{6}{*}{$320 \times 160$} & 0 & \xmark  & \xmark & \xmark  & 60  \\
                         & 1 & \xmark  & \xmark & \xmark  & 85  \\
                         & 2 & 1   & 1  & 2   & 110 \\
                         & 3 & 16  & 16 & 64  & 170 \\
                         & 4 & 64  & 16 & 96  & 215 \\
                         & 5 & 128 & 16 & 128 & 250 \\ \hline
\end{tabular}
\end{table}

\section{Details of expressivity study}
\label{express_study}

In demonstrating the effect of a neural network on the optimization landscape, we eliminated the effect of the dimensionality of the decision space, i.e., the number of trainable parameters. This was feasible since the mesh resolution was low ($64 \times 32$), allowing the network to accurately represent the baseline's designs (see Fig.~\ref{fig:app_reg_designs}), which can be assumed to be the ground truth. However, this may not always be the case, as reparameterization schemes can be over- or under-parameterized compared to the baseline. In other words, the number of decision variables (weights and biases of the neural network) can be higher or lower than the number of finite elements' density values.

The degree of over- or under-parameterization, along with the type of neural network, can affect the optimization dynamics and the network's expressivity---i.e., its ability to recreate the same design that can be represented by the finite element discretization of a chosen resolution. For instance, in standard topology optimization (TO), a density filter is applied to restrict the decision space and prevent convergence to checkerboard patterns, which are chosen by the optimizer due to their lower compliance~\cite{to_revew}.\footnote{Density filter also introduces a characteristic length-scale into the design.} While such a restriction is beneficial, in this scenario the restriction of the decision space due to neural networks could prevent convergence to certain designs for a particular mesh resolution.

Herein we use a simple strategy to assess the network expressivity by assuming that baseline solutions represent the ground truth (for which we used MMA with tuned hyperparameters). We then quantify the neural networks' ability to represent such topologies and use that quantity as a measure of the expressivity.
We use the peak signal-to-noise ratio (PSNR), which is commonly used in computer vision literature to measure the discrepancy between a noisy (or reconstructed) image and the ground truth. The PSNR is calculated as
\begin{equation}
    \text{PSNR} = 10\log_{10}\frac{\text{R}^2}{\text{MSE}},
\end{equation}
where $\text{R}$ is the maximum possible pixel value that we set to $\text{R} = 1$, and MSE refers to the mean-squared error. A high PSNR value corresponds to a design that is visually similar to that obtained by the baseline. For instance, the lowest value in Fig.~\ref{fig:app_reg_designs} is $\text{PSNR} \approx 33$ and is visually indistinguishable.

\begin{figure}
     \centering
     \resizebox{0.3\textwidth}{!}{
       \includegraphics{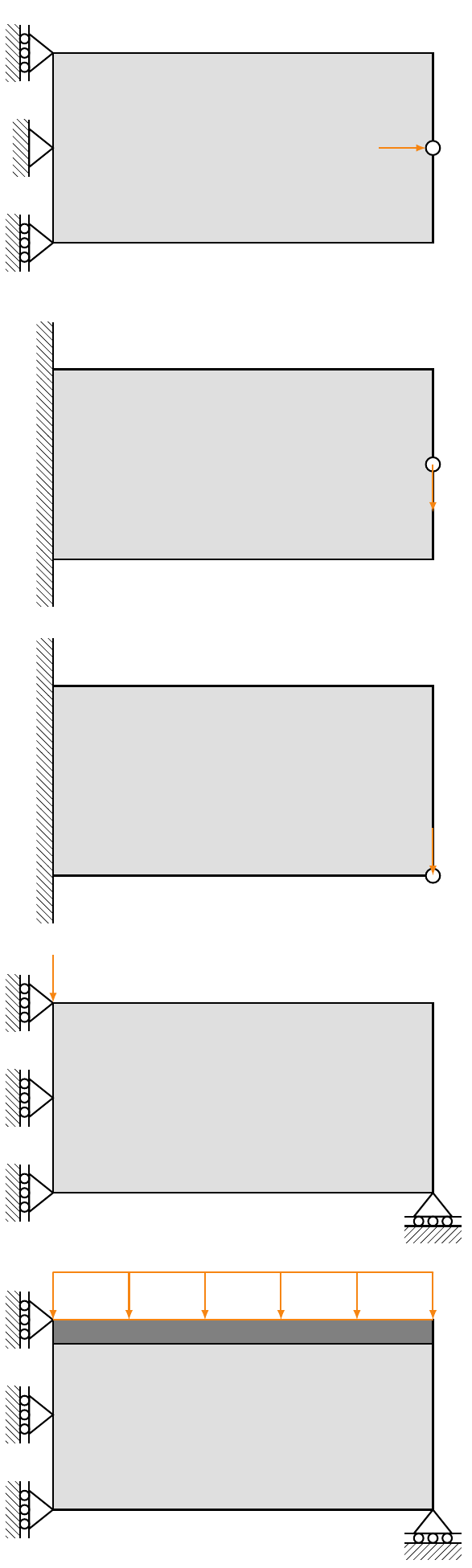}
     }
     \caption{Different boundary value problems used in this study. All loads are distributed (not concentrated) across a finite length. For the bridge case, the top few elements have been designated as non-designable and was set to have material density of 1.0.}
     \label{fig:bc}
 \end{figure}

First, we generated baseline solutions for three test cases---namely, the MBB, Michell, and cantilever beams (see Figure~\ref{fig:bc})---at the required mesh resolution, with the target volume set at  $V_0 = 30\%$. Notice that the optimized designs for these bending-dominated problems will portray fine structural features. Next, we selected a series of networks---both under- and over-parameterized---with different number of network parameters. Namely, the ratios between the network and baseline parameters are $\left\{ \frac{3}{10},  \frac{6}{10}, \frac{23}{10}, \frac{37}{10}, \frac{50}{10} \right\}$ (see Table~\ref{tab:app_exp_table} for exact architecture details)\footnote{We note that the CNN architecture cannot be under-parameterized when compared to the number of finite elements, thus all CNN networks are only over-parameterized.}.
Second, we solved Eq.~\eqref{eqn:regression_opt} for each network and test case combination by substituting $\boldsymbol{{\rho}}$ with the target baseline solution $\boldsymbol{\rho}^{\star}$ and subsequently computed its corresponding PSNR value. After obtaining the PSNR values for the three test cases, the worst value was taken as the measure for that particular network at that mesh resolution. Because this measure is influenced by the optimizer and its hyperparameters, we also conducted hyperparameter tuning to maximize the PSNR value. Finally, we repeated this procedure five times with different starting points for the hyperparameter optimization to obtain PSNR values for each set of the best-identified hyperparameters. This allowed us to calculate the mean and confidence bounds. We repeated the experiment for four different mesh resolutions ($64 \times 32$, $128 \times 64$, $256 \times 128$, and $320 \times 160$).


The results of this expressivity study are summarized in Fig.~5 of the main document. We also considered the CNN used by Hoyer et al.~\cite{hoyer}, that even though is over-parameterized by approximately \num{120} times, it achieved similar PSNR values (71, 74, 75, and 74 for all four resolutions)\footnote{Not shown in Fig.~\ref{fig:expressivity} because the number of parameters is much larger than the other CNN networks considered.}.
Regarding the MLP and SIREN network architectures, in most published works the network size is kept constant without regard to the mesh resolution~\cite{tounn, Deng2020_tounn2, Chandrasekhar2021}. A vertical dashed line in the figure indicates the PSNR results attained by a network architecture with approximately \num{2000} parameters. At the highest resolution, this fixed architecture must represent a design with $320 \times 160 = \num{51200}$ pixels (or densities) using approximately \num{2000} parameters, resulting in over a \num{250}-fold compression and degrading the representation quality accordingly. Thus, using a network of fixed size produces simpler structures as the mesh resolution is increased. Finally, the superior performance of SIREN compared to MLP can be attributed to SIREN’s ability to represent high frequencies necessary for sharp solid-void transitions, similar to square waveforms. MLP struggles with this due to its spectral bias, which causes it to fit lower-frequency signals first and capture higher frequencies very slowly.

\section{Topology optimization details}
\label{top_opt_details}

For all the boundary values problems considered in this study (see Fig.~\ref{fig:bc}), Young's moduli were set at $10^{-9}$ for the void and $10$ for the material, and Poisson's ratio was set to  $0.3$; plane stress conditions were assumed. For solving the finite element discrete system of equations, a direct LU solver was used. We use the Python implementation of MMA from~\cite{mma_python}.

\subsection{Initialization}
The baseline is initialized with a uniform density field, where each pixel is set to the target volume fraction. To be consistent with the baseline and to start the optimization from a feasible point, we trained the NNs' parameters to generate a uniform gray density distribution before starting topology optimization~\cite{Zhang2021_tonr}. This can be achieved by solving Eq.~\eqref{eqn:regression_opt}, by setting ${\rho}_i = V_0$. This pre-training was performed with $\num{300}$ iterations of Adam~\cite{kingma2014adam} (with a default learning rate of $0.001$), yielding errors lower than $10^{-4}$. Noteworthy, the cost of this operation is negligible since neither the expensive finite element analysis nor the adjoint analysis is conducted.

\subsection{Thresholding designs}
To obtain black-and-white designs---i.e., density values of either 0 or 1 for all finite elements---we use the algorithm described by Sigmund~\cite{Sigmund2022}. Briefly, if the number of elements/pixels is $N$, then the design is flattened and sorted in descending order based on the densities. Then, the number of pixels to be set to black $N_p$ is obtained by
\begin{equation}
    N_p = \frac{N \left( V_0 - 0.001 \right)}{1 - 0.001}
\end{equation}
where $V_0$ is the target volume fraction. Then, the discrete design is obtained by setting the first $N_p$ values to $1$ and all others to $0.001$~\cite{Sigmund2022}. If the volume fraction changes slightly, a new compliance value can be calculated as $c_{\text{new}} = c_{\text{th}} \times \frac{V_{0, \text{th}}}{V_0}$, where $c_{\text{th}}$ is the compliance value of the thresholded design with a volume fraction $V_{\text{th}}$, which (slightly) violates the target volume fraction $V_0$.

 \begin{figure*}
    \centering
    	\includegraphics{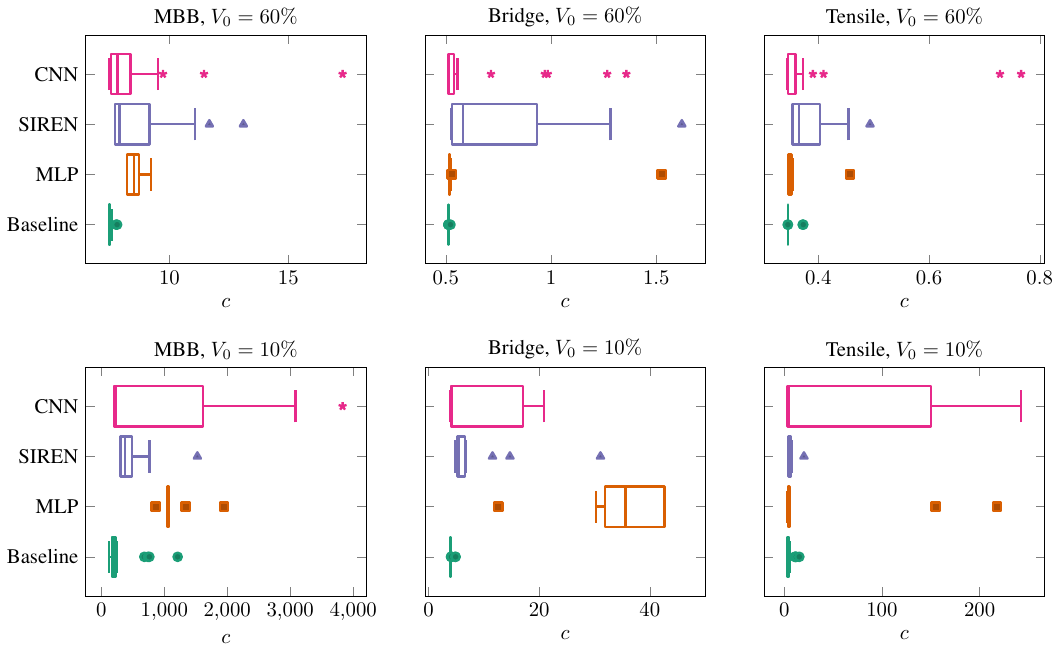}
    \caption{Robustness to hyperparameter selection when using the MMA optimizer. The box plot shows the variation in the compliance $c$ as hyperparameters are varied for the different models. The outliers are marked while the median (middle of the box) and the 25\% and 75\% quartiles are shown (box's ends).}
    \label{fig:hp_mma}
\end{figure*}

 \begin{figure*}
    \centering
    	\includegraphics{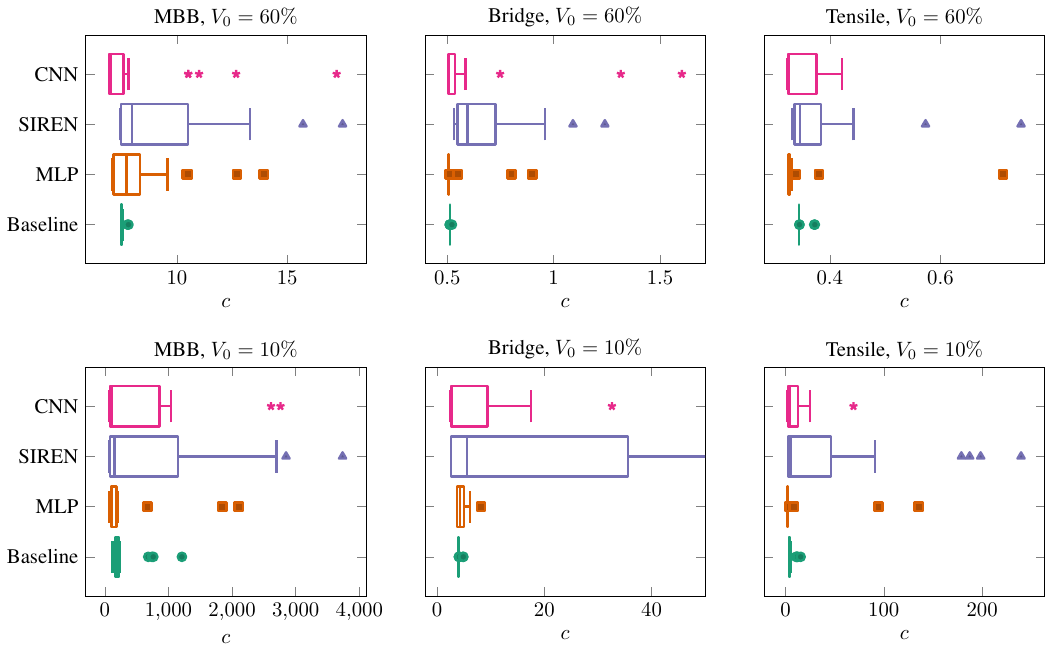}
    \caption{Robustness to hyperparameter selection when Adam is used for optimizing the networks. The box plot shows the variation in the compliance $c$ as hyperparameters are varied for the different models. The outliers are marked while the median (middle of the box) and the 25\% and 75\% quartiles are shown (box's ends). }
    \label{fig:hp_adam}
\end{figure*}

\subsection{Hyperparameter tuning}

Since the performance of optimizers hinges on the hyperparameters employed, they were individually tuned for each test case across all problems, including both the baseline and the neural reparameterizations, to ensure optimal performance across all methods. Tuning these hyperparameters can be costly since they involve a two-level optimization, i.e., topology optimization is conducted for each set of hyperparameters.Therefore, we tuned the hyperparameters only at the lowest resolution ($64 \times 32$) by performing $60$ iterations with the optimizer, and used the hyperparameters that minimized the compliance for the optimization at higher resolutions. The hyperparameters chosen for the different methods were:
\begin{enumerate}    
\item MMA
    \begin{itemize}
        \item Move limit $m$;
        \item Asymptote initialization $a$;
        \item Bounds on decision variables $b$ (only for training neural networks)
    \end{itemize}
\item Adam
    \begin{itemize}
        \item Learning rate $\eta$;
        \item Gradient clipping value $g_c$ (i.e., scaling of the gradient vector if its norm is above this value).
    \end{itemize}    
\end{enumerate}

\begin{table*}[t]
\centering
\caption{Best hyperparameters found for the problems (tensile and Michell cases with 60\% material) considered in landscape analysis. For MMA, the hyperparameters are ordered as move limit $m$, asyinit $a$, and variable bound $b$ while for Adam, the learning rate $\eta$ and gradient clipping value $g_c$ are shown. For SIREN, the $\omega_0$ values are also shown as the last entry.}
\label{tab:hyper_toy}
\begin{tabular}{@{}cccccc@{}}
\toprule
\textbf{Optimizer} & \textbf{Problem} & \textbf{Baseline} & \textbf{MLP} & \textbf{SIREN} & \textbf{CNN} \\ \midrule
\multicolumn{2}{c}{\textbf{Hyperparameters:}} & $m,\ a$ & $m,\ a,\ b$ & $m,\ a,\ b,\ \omega_0$ & $m,\ a,\ b$ \\ \midrule
\multirow{4}{*}{MMA} & Tensile ($p=1$) & $0.03, 0.5$ & $0.03, 0.4, 11$ & $2 \times 10^{-4}, 0.1, 5, 25$ & $0.056, 0.3, 11$ \\
& Michell ($p=1$) & $0.056, 0.3$ & $0.002, 0.4, 8$ & $0.001, 0.1, 2, 10$ & $0.0056, 0.2, 5$ \\
& Tensile ($p=3$) & $0.1, 0.2$ & $0.003, 0.4, 5$ & $0.002, 0.2, 2, 5$ & $0.0056, 0.5, 5$ \\ 
& Michell ($p=3$) & $0.1, 0.2$ & $0.003, 0.2, 2$ & $0.002, 0.2, 2, 10$ & $0.003, 0.1, 2$ \\ \midrule
\multicolumn{2}{c}{\textbf{Hyperparameters:}} &  & $\eta,\ g_c$ & $\eta, \ g_c,\ \omega_0$ & $\eta, \ g_c$ \\ \midrule
\multirow{4}{*}{Adam} & Tensile ($p=1$) & \xmark & $0.02, 1 \times 10^{-4}$ & $0.01, 0.01, 5$ & $0.03, 0.01$ \\
& Michell ($p=1$) & \xmark & $0.056, 0.1$ & $0.0056, 0.1, 15$ & $0.056, 1 \times 10^{-4}$ \\
& Tensile ($p=3$) & \xmark & $0.02, 0.1$ & $0.0056, 0.1, 15$ & $0.03, 0.01$ \\
& Michell ($p=3$) & \xmark & $0.03, 1 \times 10^{-4}$ & $0.01, 1 \times 10^{-4}, 15$ & $0.03, 0.01$ \\ \bottomrule
\end{tabular}
\end{table*}

In addition to these optimizer-specific hyperparameters, SIREN has an additional hyperparameter $\omega_0$, which controls the frequencies that can be learned. All hyperparameters were tuned with the Optuna package~\cite{akiba2019optuna} using the tree-structured Parzen estimator (TPE) algorithm. For the test cases used in the landscape visualization, the best hyperparameter values obtained after $25$ outer iterations are given in Table \ref{tab:hyper_toy}. Further, for the MBB, bridge, and tensile test cases considered for benchmarking, a study on robustness to hyperparameters is shown in Figs.~\ref{fig:hp_mma} and~\ref{fig:hp_adam} using box plots; these show the median, 25\% and 75\% quartiles, as well as some outliers. These results were taken from the hyperparameter optimization performed before the benchmarking, after removing the hyperparameters that caused divergence. 

\subsection{Performance profiles}
\label{perf_prof}

Performance profiles \cite{Dolan2002} are used for statistically comparing $s$ solvers (or methods) on $t$ test cases (see \cite{RojasLabanda2015} for an application to TO). To construct a performance profile, the performance ratio for solver $i$ on case $j$ is defined as
\begin{equation*}
    r_{ij} = \frac{M_{ij}}{\min(M_{1j}, M_{2j}, \cdots, M_{nj})}, \qquad M_{ij}\geq 0,
\end{equation*}
 where $M$ is any scalar metric of interest (e.g., best objective value, converged iteration, or compliance of the thresholded design) that is to be compared.  This ratio indicates the performance relative to the best solver for that particular case. Next,  the solver is treated as a ``winner'' for a particular case if its performance is within a tolerance of the best solver according to the following function:
 \begin{equation}
     k(r_{ij}, \tau) = \begin{cases}
                        1 & r_{ij} \leq \tau, \\
                        0 & \text{otherwise},
                        \end{cases}
 \end{equation}
where $\tau \geq 1$ is the tolerance factor.  If $\tau=1$, the allowable error is 0\%, and only one solver is allowed to be the winner for a given test case. As we relax the tolerance from $1 \rightarrow \infty$, more solvers are considered winners. The performance profile is then the evolution of the percentage of test cases where the solver is a winner as the tolerance is relaxed. Thus, the performance profile for the $i$th solver is given by
\begin{equation}
    p_i(\tau) = \frac{\sum_j k(r_{ij}, \tau)}{m},
\end{equation}
which denotes the probability that the solver's performance is within a factor $\tau$ of the best possible performance for all test cases. For all plots showing performance profiles, the allowed tolerance (in \%) is used as the abscissas instead of $\tau$. 

\begin{figure*}
    \centering
    \includegraphics{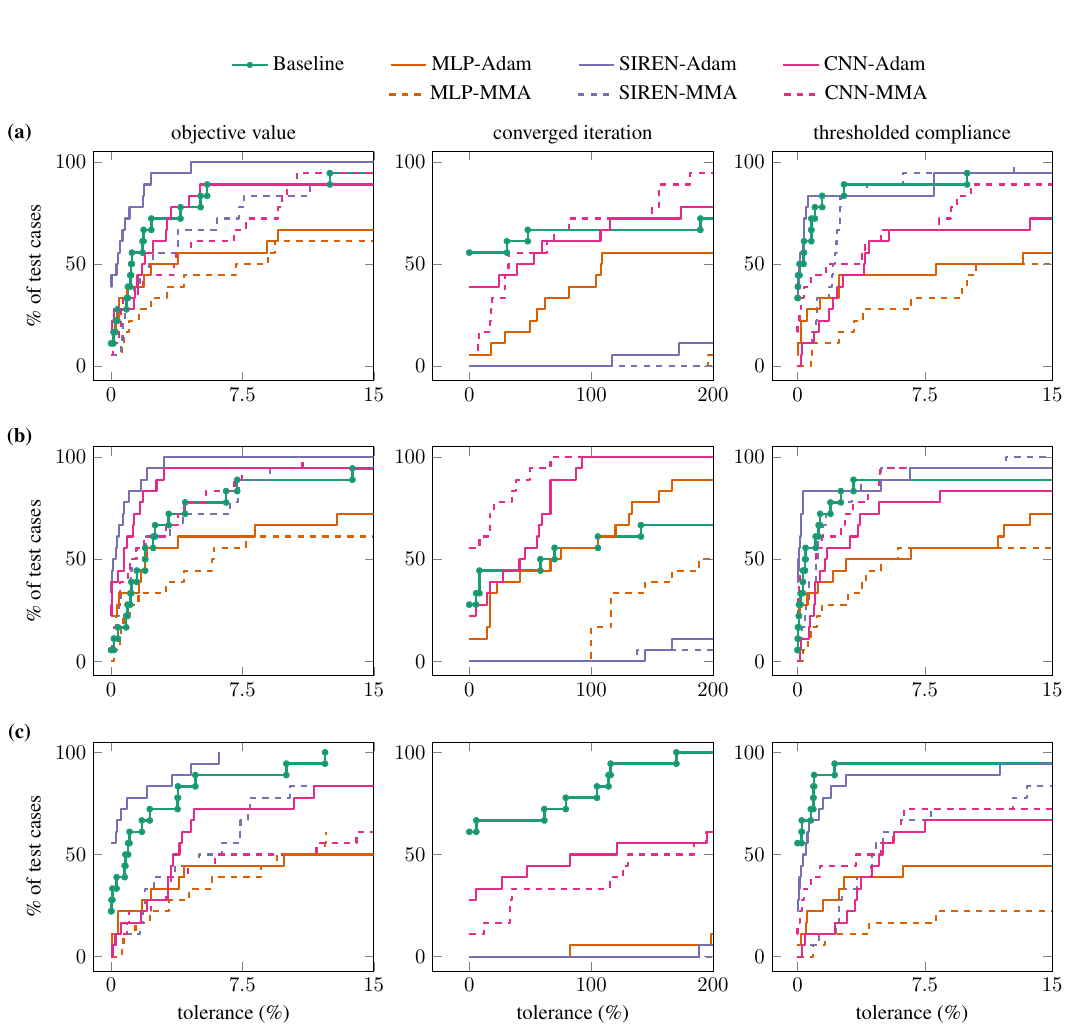}
    \caption{Performance profiles showing \textbf{(a)} Median performance, \textbf{(b)} best performance, and \textbf{(c)} worst performance across six different network initializations. For all methods, the profiles compare the best objective value attained during optimization (left column), the number of iterations to convergence (middle column), and the compliance after thresholding to black-and-white designs (right column). The plot shows the percentage of test cases (ordinates) where each method achieved results within a tolerance (abscissas) of the best-performing method for that case. Neural TO was carried out using both Adam and MMA optimizers, while the baseline method only used MMA. All networks have parameters corresponding to the a mesh resolution of $576 \times 288$ elements, and were pretrained to start with a uniform density distribution.}
    \label{fig:overall_pp_si}
\end{figure*}

\begin{figure*}
    \centering
    	\includegraphics{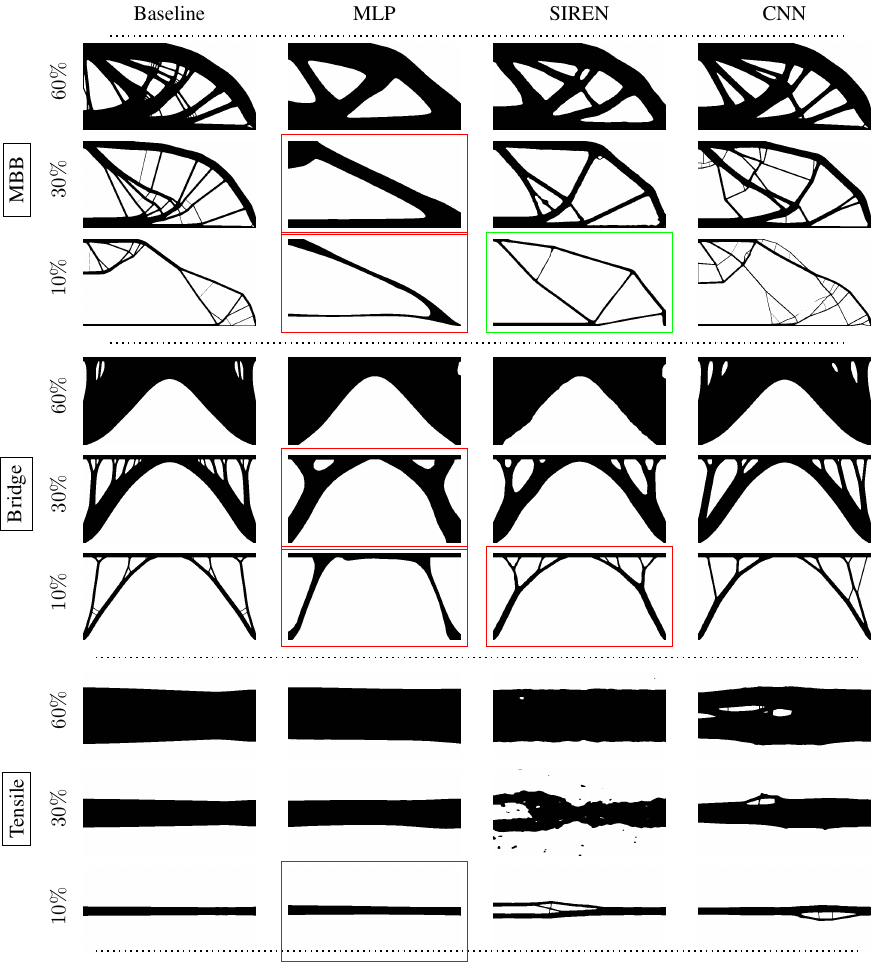}
    \caption{Typical thresholded designs obtained for the three test cases--MBB, Bridge, and Tensile cases for $V_0$ = 60\%, 30\%, and 10\% when optimized with MMA.}
    \label{fig:overall_designs}
\end{figure*}

\begin{figure*}
    \centering
    	\includegraphics{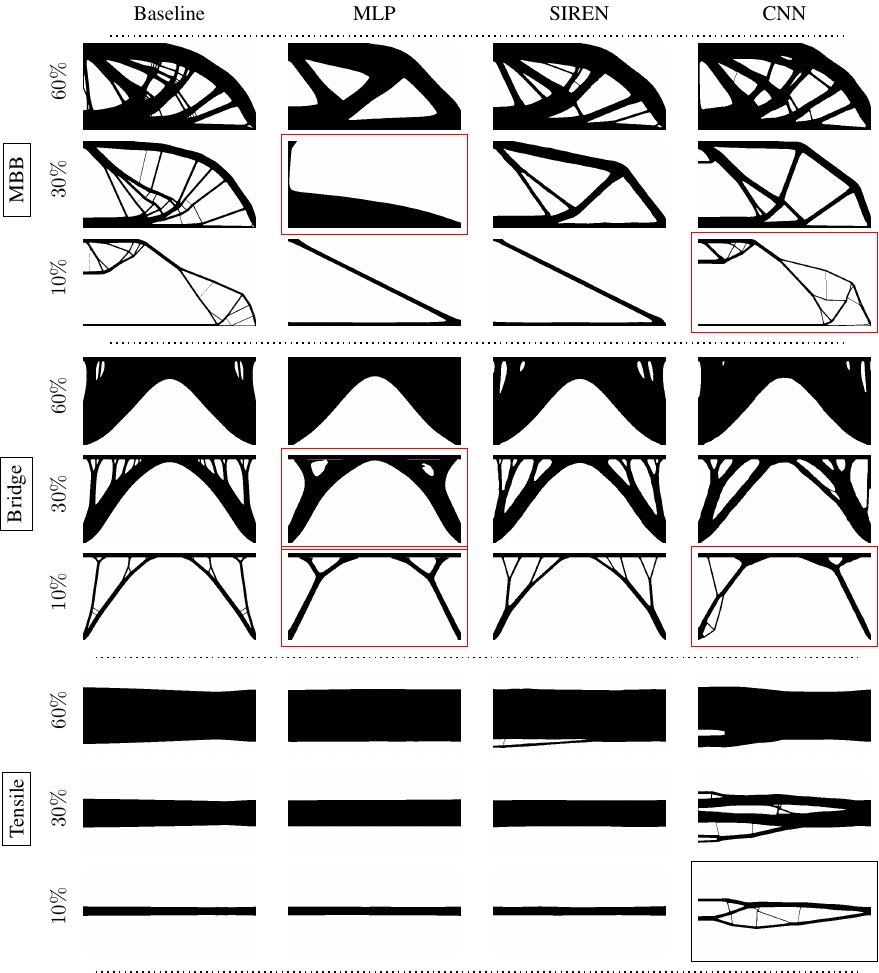}
    \caption{Typical thresholded designs obtained for the three test cases--MBB, Bridge, and Tensile cases for $V_0$ = 60\%, 30\%, and 10\% when optimized with Adam.}
    \label{fig:overall_designs_adam}
\end{figure*}

In Fig.~\ref{fig:overall_pp_si} we compare the performances using median, best, and worst values obtained across six different network initializations. The raw values of the metrics scaled to the value of the baseline are given in Tables~\ref{tab:mma} and~\ref{tab:adam}. Typical designs obtained after thresholding are shown in Figs.~\ref{fig:overall_designs} and~\ref{fig:overall_designs_adam} for the three test cases at target volume fractions of 10\%, 30\%, and 60\%. Most designs shown have compliance values similar to the baseline, except for those marked in red (which are worse by at least 10\%). The designs marked in green have lower compliance values. As expected, MLP's designs are simple and lack fine features. In contrast, CNNs produce designs with finer features due to their high expressivity, even for the tensile case.

\subsection{Additional results}
\label{add_results}

In addition to compliance minimization, we explored the effectiveness of neural topology TO for two other problems: thermal conduction and compliant mechanism design, following Wang et al.~\cite{Wang2010}. Both problems follow a formulation similar to that of compliance minimization, but differ in their specific objective functions and physical parameters.

For the thermal conduction problem, the goal is to optimize the material distribution to efficiently dissipate heat from a design domain to a sink. The material conductivity is set to $1.0$, while the void or non-material regions have a conductivity of $0.001$. The objective function $\mathcal{F}$ is expressed as $\mathcal{F} = \boldsymbol{P}^\intercal \boldsymbol{U}$, where $\boldsymbol{P}=\boldsymbol{F}$ represents the unit thermal load distributed throughout the domain. For this problem we set the target volume to $V_0 = 30\%$.

For the compliant mechanism design problem, the objective is to construct a mechanism where the force applied at the top-left node of the domain yields the maximum negative displacement at the top-right node. Here, the input spring stiffness ($k_{in}$) and input force are both set to $1$, while the output spring stiffness ($k_{out}$) is set to 0.001. All entries in vector $\boldsymbol{P}$ are zeros except for the one corresponding to the output degree of freedom (the top-right node), which is set to one. The target volume is set to $V_0 = 40\%$.

\begin{figure*}
    \centering
    	\includegraphics{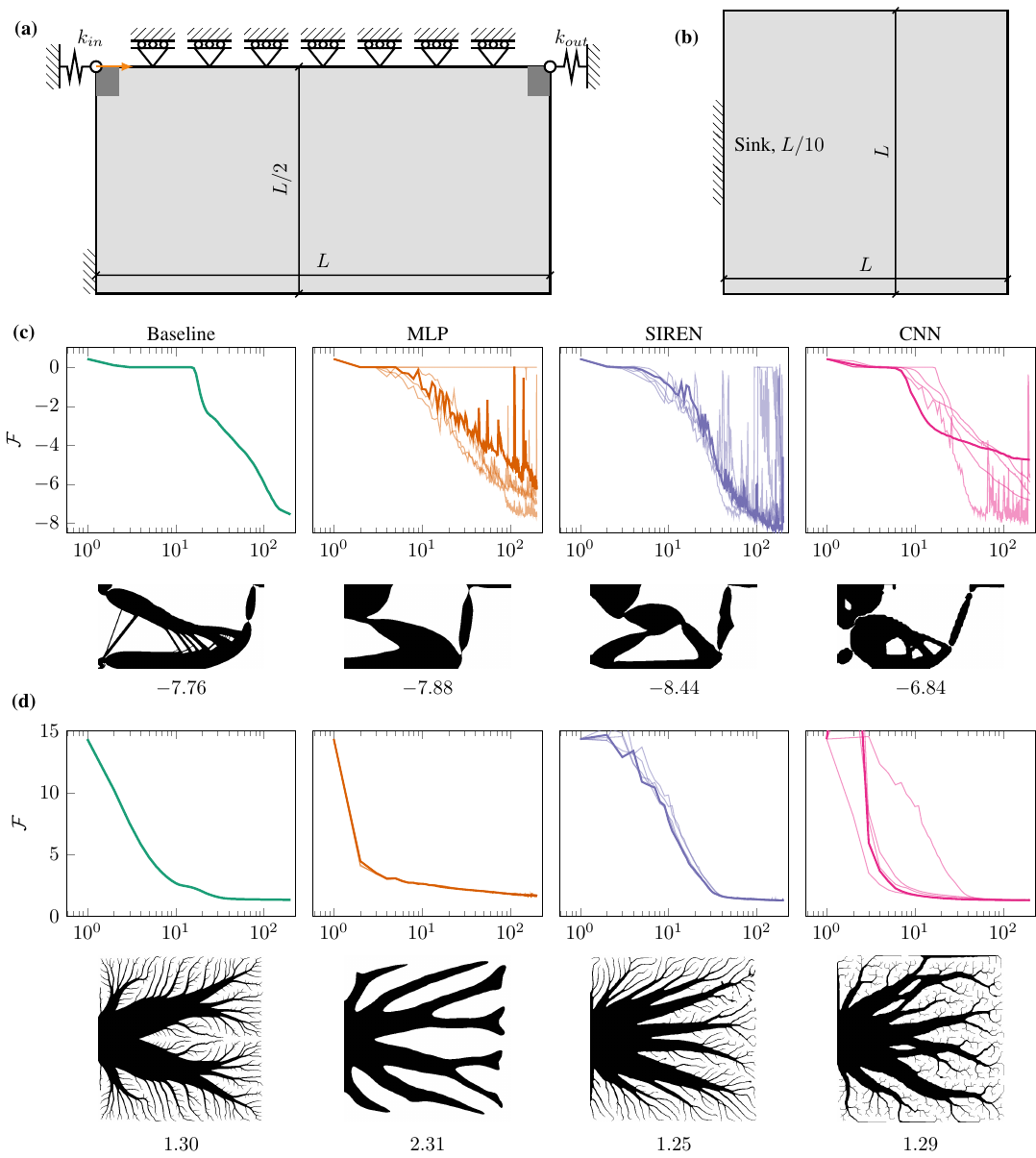}
    \caption{Comparing neural TO's performance against baseline on compliant mechanism design \textbf{(a)} and thermal conduction \textbf{(b)} problems, when optimized with Adam. For the neural networks, the curves have been repeated 6 times (each time with a different starting point). Note that all networks are pre-trained to generate uniform gray density field. The plots show the objective value's evolution with optimization iteration. The best thresholded designs and their objective values are also shown below the corresponding scheme. }
    \label{fig:thermal_mech}
\end{figure*}

We follow the same procedure as with compliance minimization, i.e., pretraining to uniform initialization, hyperparameter tuning at $64\times32$ resolution and testing at $544\times272$.\footnote{Since the thermal problem has a square domain, we use the same number of elements for x and y directions i.e. we test at $544\times544$ mesh resolution} The results, as well as schematics of the boundary conditions, are shown in Fig.~\ref{fig:thermal_mech}. We only used Adam as the optimizer since it was better than MMA for the compliance minimization problem. It can be seen that MLP still produces designs that lack fine features which both CNN and SIREN can create. However, for the mechanism design problem, both MLP and SIREN can get designs that outperform the baseline. However, we see that the objective function curves for the networks are non-smooth, unlike the baseline.
 
\begin{sidewaystable}
\centering
\caption{Values of the three metrics used for constructing the performance profiles after optimizing with MMA. These values have been scaled against the baseline. Thus, any value $<1$ can be considered to beat the baseline.}
\label{tab:mma}
\begin{tabular}{@{}c|lll|lll|lll@{}}
\toprule
 & \multicolumn{3}{c}{\textbf{Objective value}} & \multicolumn{3}{c}{\textbf{Converged iteration}} & \multicolumn{3}{c}{\textbf{Thresholded compliance}} \\ \cmidrule(l){2-10} 
  $V_0$ & 
  CNN &
  SIREN &
  MLP &
  CNN &
  SIREN &
  MLP &
  CNN &
  SIREN &
  MLP \\ \midrule
\rowcolor{lightgray!50} \multicolumn{10}{c}{\textbf{MBB}}  \\
$0.6$ &
  $0.99$, $0.99$, $0.99$ &
  $0.99$, $0.99$, $1.01$ &
  $1.03$, $1.03$, $1.03$ &
  $0.41$, $0.16$, $0.65$ &
  $2.05$, $1.44$, $2.58$ &
  $0.93$, $0.33$, $3.79$ &
  $0.99$, $0.99$, $0.99$ &
  $1.01$, $1.0$, \xmark &
  $1.03$, $1.03$, $1.03$ \\
$0.5$ &
  $1.0$, $1.0$, $1.0$ &
  $1.01$, $1.0$, $1.02$ &
  $1.01$, $1.01$, $1.01$ &
  $2.81$, $1.5$, $3.88$ &
  $14.38$, $10.5$, $19.38$ &
  $9.19$, $6.75$, $18.12$ &
  $1.0$, $1.0$, $1.0$ &
  $1.02$, $1.02$, $1.04$ &
  $1.01$, $1.01$, $1.01$ \\
$0.4$ &
  $1.03$, $1.0$, \xmark &
  $0.99$, $0.99$, $0.99$ &
  $1.0$, $0.99$, $1.11$ &
  $2.56$, $1.35$, $3.76$ &
  $7.09$, $3.53$, $9.65$ &
  $8.29$, $4.35$, $11.24$ &
  $1.1$, $1.02$, $1.18$ &
  $1.0$, $1.0$, $1.01$ &
  $1.03$, $1.01$, $1.15$ \\
$0.3$ &
  $0.99$, $0.98$, $1.03$ &
  $1.0$, $0.99$, $1.02$ &
  $1.06$, $1.03$, $1.09$ &
  $0.37$, $0.11$, $0.52$ &
  $1.69$, $1.24$, $2.52$ &
  $0.98$, $0.24$, $2.61$ &
  $0.99$, $0.98$, $0.99$ &
  $1.01$, $1.0$, $1.01$ &
  $1.05$, $1.03$, $1.07$ \\
$0.2$ &
  $1.0$, $1.0$, $1.01$ &
  $1.02$, $1.01$, $1.02$ &
  $1.03$, $1.03$, $1.03$ &
  $2.56$, $1.67$, $12.78$ &
  $16.56$, $3.22$, $17.78$ &
  $9.11$, $5.78$, $16.22$ &
  $1.0$, $1.0$, $1.01$ &
  $1.02$, $1.01$, $1.03$ &
  $1.1$, $1.03$, $2.09$ \\
$0.1$ &
  $1.0$, $0.99$, $1.06$ &
  $1.0$, $0.99$, $1.02$ &
  $1.0$, $1.0$, $1.02$ &
  $0.91$, $0.42$, $2.3$ &
  $2.47$, $1.74$, $3.11$ &
  $3.15$, $0.83$, $3.66$ &
  $1.0$, $1.0$, $1.0$ &
  $1.01$, $1.01$, $1.02$ &
  $1.02$, $1.01$, $1.22$ \\
\midrule
\rowcolor{lightgray!50} \multicolumn{10}{c}{\textbf{Bridge}}  \\
$0.6$ &
  $1.04$, $1.01$, $1.12$ &
  $1.04$, $1.01$, $1.08$ &
  $1.22$, $1.22$, $1.23$ &
  $0.43$, $0.16$, $4.65$ &
  $2.33$, $1.12$, $2.91$ &
  $1.49$, $0.33$, $3.35$ &
  $1.05$, $1.02$, $1.05$ &
  $1.06$, $1.05$, $1.16$ &
  $1.23$, $1.21$, $1.24$ \\
$0.5$ &
  $1.0$, $1.0$, $1.0$ &
  $1.02$, $1.01$, $1.03$ &
  $1.08$, $1.07$, $1.09$ &
  $1.69$, $0.92$, $3.38$ &
  $8.92$, $3.69$, $10.54$ &
  $7.65$, $4.46$, $10.62$ &
  $1.0$, $1.0$, $1.0$ &
  $1.02$, $1.01$, $1.03$ &
  $1.1$, $1.06$, $4.98$ \\
$0.4$ &
  $1.01$, $0.99$, $1.03$ &
  $1.0$, $0.99$, $1.0$ &
  $1.0$, $0.99$, $1.01$ &
  $3.12$, $1.38$, $6.23$ &
  $9.19$, $5.38$, $13.31$ &
  $13.04$, $4.38$, $15.0$ &
  $1.0$, $1.0$, $1.0$ &
  $1.01$, $1.0$, $1.02$ &
  $1.01$, $1.0$, $1.02$ \\
$0.3$ &
  $1.02$, $1.0$, $1.12$ &
  $1.01$, $1.0$, $1.03$ &
  $1.25$, $1.25$, $1.26$ &
  $0.34$, $0.16$, $0.62$ &
  $1.83$, $0.96$, $2.18$ &
  $1.41$, $0.29$, $2.27$ &
  $1.03$, $1.02$, $1.03$ &
  $1.02$, $1.01$, $1.06$ &
  $1.27$, $1.26$, $1.27$ \\
$0.2$ &
  $1.0$, $0.99$, $1.01$ &
  $1.02$, $1.01$, $1.04$ &
  $1.19$, $1.17$, $1.21$ &
  $1.32$, $0.71$, $2.94$ &
  $5.91$, $3.53$, $8.59$ &
  $4.82$, $2.53$, $6.59$ &
  $1.0$, $1.0$, $1.01$ &
  $1.02$, $1.01$, $1.8$ &
  $1.16$, $1.15$, $1.18$ \\
$0.1$ &
  $1.08$, $1.03$, $1.19$ &
  $1.41$, $1.05$, $1.77$ &
  $1.0$, $0.99$, $1.0$ &
  $1.82$, $0.95$, $2.84$ &
  $6.71$, $3.16$, $8.11$ &
  $5.68$, $2.32$, $9.05$ &
  $1.1$, $1.04$, $1.18$ &
  $0.53$, $0.53$, $0.53$ &
  $1.02$, $1.01$, $1.07$ \\
\midrule
\rowcolor{lightgray!50} \multicolumn{10}{c}{\textbf{Tensile}}  \\
$0.6$ &
  $1.07$, $1.01$, $1.17$ &
  $1.01$, $0.97$, $1.05$ &
  $1.38$, $1.32$, $1.47$ &
  $0.4$, $0.14$, $0.62$ &
  $2.36$, $1.19$, $3.19$ &
  $1.06$, $0.38$, $2.55$ &
  $1.09$, $1.04$, $1.17$ &
  $1.02$, $0.99$, $1.05$ &
  $1.33$, $1.3$, $1.35$ \\
$0.5$ &
  $1.0$, $0.99$, $1.01$ &
  $1.03$, $1.02$, $1.03$ &
  $1.59$, $1.53$, $1.63$ &
  $1.08$, $0.68$, $2.26$ &
  $6.74$, $2.11$, $7.0$ &
  $5.11$, $2.58$, $7.42$ &
  $1.01$, $1.0$, $1.05$ &
  $1.03$, $1.02$, $1.03$ &
  $1.36$, $1.34$, $1.42$ \\
$0.4$ &
  $1.07$, $0.99$, $1.32$ &
  $0.99$, $0.99$, $1.05$ &
  $1.0$, $0.99$, $1.03$ &
  $2.5$, $1.0$, $4.13$ &
  $8.43$, $6.67$, $11.13$ &
  $9.83$, $4.33$, $13.07$ &
  $1.08$, $1.05$, $1.24$ &
  $1.01$, $1.0$, $1.08$ &
  $1.11$, $1.01$, $1.34$ \\
$0.3$ &
  $0.95$, $0.88$, $1.04$ &
  $0.96$, $0.9$, $1.06$ &
  $8.93$, $4.04$, \xmark &
  $0.37$, $0.11$, $0.5$ &
  $1.69$, $0.93$, $1.98$ &
  $0.98$, $0.24$, $3.06$ &
  $0.99$, $0.86$, $1.05$ &
  $0.91$, $0.87$, $1.12$ &
  \xmark, $1.78$, \xmark \\
$0.2$ &
  $0.98$, $0.97$, $1.07$ &
  $1.04$, $1.03$, $1.05$ &
  $5.14$, $4.69$, $7.5$ &
  $0.68$, $0.49$, $0.95$ &
  $3.28$, $2.19$, $4.7$ &
  $2.11$, $1.41$, $3.81$ &
  $1.01$, $1.0$, $1.03$ &
  $1.1$, $1.09$, $1.11$ &
  $5.49$, $4.42$, $12.72$ \\
$0.1$ &
  $1.1$, $0.98$, $1.23$ &
  $1.02$, $0.98$, $1.05$ &
  $1.04$, $0.99$, $1.07$ &
  $1.63$, $1.08$, $2.15$ &
  $4.5$, $2.19$, $6.31$ &
  $4.79$, $3.23$, $6.46$ &
  $1.16$, $1.01$, $1.26$ &
  $1.03$, $1.0$, $1.05$ &
  $1.22$, $1.04$, $1.84$ \\ \bottomrule
\end{tabular}
\end{sidewaystable}

\begin{sidewaystable}
\centering
\caption{Values of the three metrics used for constructing the performance profiles after optimizing the networks with Adam optimizer. These values have been scaled against the baseline (optimized with Adam). Thus, any value $<1$ can be considered to beat the baseline.}
\label{tab:adam}
\begin{tabular}{@{}c|lll|lll|lll@{}}
\toprule
 & \multicolumn{3}{c}{\textbf{Objective value}} & \multicolumn{3}{c}{\textbf{Converged iteration}} & \multicolumn{3}{c}{\textbf{Thresholded compliance}} \\ \cmidrule(l){2-10} 
  $V_0$ & 
  CNN &
  SIREN &
  MLP &
  CNN &
  SIREN &
  MLP &
  CNN &
  SIREN &
  MLP \\ \midrule
\rowcolor{lightgray!50} \multicolumn{10}{c}{\textbf{MBB}}  \\
$0.6$ &
  $0.99$, $0.99$, $1.01$ &
  $1.0$, $0.99$, $1.0$ &
  $1.01$, $1.0$, $1.01$ &
  $0.31$, $0.26$, $0.49$ &
  $1.67$, $0.91$, $3.14$ &
  $0.57$, $0.16$, $4.65$ &
  $1.0$, $1.0$, $1.02$ &
  $0.99$, $0.99$, $1.0$ &
  $1.02$, $1.01$, $1.02$ \\
$0.5$ &
  $1.0$, $1.0$, $1.0$ &
  $1.0$, $1.0$, $1.0$ &
  $1.0$, $1.0$, $1.0$ &
  $3.44$, $1.88$, $4.62$ &
  $8.25$, $5.12$, $17.0$ &
  $5.19$, $3.75$, \xmark &
  $1.0$, $1.0$, $1.0$ &
  $1.0$, $1.0$, $1.0$ &
  $1.01$, $1.01$, $1.01$ \\
$0.4$ &
  $1.0$, $1.0$, $1.03$ &
  $0.99$, $0.99$, $1.0$ &
  $0.99$, $0.99$, $0.99$ &
  $2.74$, $1.41$, $6.94$ &
  $5.79$, $3.06$, $7.59$ &
  $5.82$, $2.06$, $9.29$ &
  $1.01$, $1.0$, $1.03$ &
  $1.0$, $1.0$, $1.01$ &
  $1.0$, $1.0$, $1.0$ \\
$0.3$ &
  $0.99$, $0.98$, $1.01$ &
  $1.01$, $1.0$, $1.02$ &
  $1.0$, $1.0$, $1.04$ &
  $0.31$, $0.19$, $0.46$ &
  $1.26$, $0.8$, $2.72$ &
  $0.49$, $0.13$, $3.7$ &
  $1.01$, $0.99$, $1.02$ &
  $0.99$, $0.98$, $1.0$ &
  $1.01$, $1.01$, $1.05$ \\
$0.2$ &
  $1.0$, $1.0$, $1.0$ &
  $1.0$, $1.0$, $1.0$ &
  $1.02$, $1.01$, $1.02$ &
  $3.17$, $1.56$, $4.67$ &
  $7.11$, $4.67$, $15.0$ &
  $5.22$, $3.11$, $8.33$ &
  $1.0$, $1.0$, $1.0$ &
  $1.0$, $1.0$, $1.0$ &
  $1.01$, $1.01$, $1.01$ \\
$0.1$ &
  $1.03$, $1.01$, $1.04$ &
  $1.0$, $0.99$, $1.0$ &
  $1.32$, $0.99$, $4.57$ &
  $0.76$, $0.6$, $1.47$ &
  $1.66$, $1.32$, $2.89$ &
  $1.56$, $0.51$, $2.98$ &
  $1.04$, $1.02$, $1.05$ &
  $1.0$, $1.0$, $1.01$ &
  $1.33$, $1.0$, $4.67$ \\
  \midrule
\rowcolor{lightgray!50} \multicolumn{10}{c}{\textbf{Bridge}} \\
$0.6$ &
  $1.01$, $1.0$, $1.05$ &
  $1.02$, $1.0$, $1.03$ &
  $1.18$, $1.18$, $1.2$ &
  $0.33$, $0.19$, $0.56$ &
  $1.69$, $0.93$, $3.23$ &
  $0.38$, $0.19$, $1.88$ &
  $1.04$, $1.03$, $1.08$ &
  $1.0$, $1.0$, $1.02$ &
  $1.21$, $1.21$, $1.23$ \\
$0.5$ &
  $1.01$, $1.0$, $1.01$ &
  $0.99$, $0.99$, $0.99$ &
  $1.03$, $1.03$, $1.03$ &
  $2.08$, $0.92$, $3.38$ &
  $4.65$, $3.08$, $9.62$ &
  $3.5$, $2.15$, $15.38$ &
  $1.02$, $1.01$, $1.02$ &
  $1.0$, $1.0$, $1.0$ &
  $1.19$, $1.12$, $1.24$ \\
$0.4$ &
  $1.01$, $0.99$, $1.03$ &
  $1.0$, $0.99$, $1.0$ &
  $0.99$, $0.99$, $7.54$ &
  $3.15$, $1.92$, $5.08$ &
  $7.73$, $4.0$, $9.69$ &
  $9.0$, $2.31$, $13.23$ &
  $1.02$, $1.0$, $1.04$ &
  $1.0$, $1.0$, $1.01$ &
  $1.0$, $1.0$, $9.7$ \\
$0.3$ &
  $0.98$, $0.96$, $1.0$ &
  $0.95$, $0.93$, $0.96$ &
  $1.18$, $1.17$, $7.04$ &
  $0.32$, $0.16$, $0.47$ &
  $1.7$, $0.84$, $3.27$ &
  $0.52$, $0.13$, $3.67$ &
  $1.03$, $1.01$, $1.05$ &
  $0.99$, $0.98$, $1.01$ &
  $1.24$, $1.22$, $1.24$ \\
$0.2$ &
  $1.01$, $0.99$, $1.01$ &
  $0.98$, $0.98$, $0.98$ &
  $1.08$, $1.06$, $1.08$ &
  $1.53$, $0.59$, $1.82$ &
  $3.94$, $2.65$, $7.71$ &
  $2.09$, $1.29$, $4.65$ &
  $1.02$, $1.01$, $1.03$ &
  $1.0$, $1.0$, $1.0$ &
  $1.27$, $1.06$, $1.31$ \\
$0.1$ &
  $1.01$, $0.99$, $1.03$ &
  $0.99$, $0.99$, $0.99$ &
  $0.98$, $0.98$, $14.24$ &
  $2.16$, $1.21$, $2.95$ &
  $5.21$, $3.68$, $6.11$ &
  $5.34$, $2.42$, $9.26$ &
  $1.04$, $1.01$, $1.05$ &
  $1.0$, $1.0$, $1.0$ &
  $1.0$, $1.0$, \xmark \\
  \midrule
\rowcolor{lightgray!50} \multicolumn{10}{c}{\textbf{Tensile}}  \\
 $0.6$ &
  $1.02, 0.99, 1.1$ &
  $1.02, 1.0, 1.06$ &
  $1.16, 1.15, 2.07$ &
  $0.43, 0.24, 0.79$ &
  $1.58, 0.95, 2.83$ &
  $0.35, 0.17, 2.9$ &
  $1.14, 1.07, 1.18$ &
  $1.08, 1.05, 1.12$ &
  $1.24, 1.23,$ \xmark \\
$0.5$ &
  $1.0, 0.98, 1.01$ &
  $0.96, 0.96, 0.96$ &
  $1.17, 1.15, 1.19$ &
  $1.39, 0.63, 2.21$ &
  $3.16, 2.63, 7.74$ &
  $2.08, 0.89, 3.63$ &
  $1.04, 1.02, 1.05$ &
  $0.99, 0.99, 0.99$ &
  $1.12, 1.1, 1.14$ \\
$0.4$ &
  $1.14, 0.99, 1.25$ &
  $0.99, 0.98, 1.02$ &
  $0.98, 0.98, 0.98$ &
  $3.77, 1.6, 6.33$ &
  $5.5, 4.13, 7.4$ &
  $9.33, 2.67, 10.73$ &
  $1.26, 1.02, 1.44$ &
  $1.01, 1.0, 1.03$ &
  $1.0, 1.0, 1.03$ \\
$0.3$ &
  $0.87, 0.82, 1.01$ &
  $0.91, 0.84, 0.91$ &
  $0.94, 0.93, 1.53$ &
  $0.29, 0.19, 0.37$ &
  $1.19, 0.65, 2.26$ &
  $0.37, 0.13, 1.48$ &
  $1.08, 1.02, 1.7$ &
  $0.98, 0.9, 0.99$ &
  $0.98, 0.98, 0.99$ \\
$0.2$ &
  $1.12, 1.05, 1.22$ &
  $0.89, 0.88, 0.89$ &
  $1.38, 1.34, 1.59$ &
  $0.68, 0.51, 1.0$ &
  $1.84, 1.19, 3.65$ &
  $1.01, 0.81, 1.73$ &
  $1.34, 1.19, 1.6$ &
  $0.97, 0.97, 0.98$ &
  $1.36, 1.32, 1.58$ \\
$0.1$ &
  $1.0, 0.94, 1.14$ &
  $0.95, 0.95, 0.95$ &
  $0.96, 0.95, 0.96$ &
  $1.6, 0.92, 3.08$ &
  $3.92, 2.46, 4.69$ &
  $4.0, 1.62, 7.19$ &
  $1.16, 1.04,$ \xmark &
  $1.0, 1.0, 1.0$ &
  $1.0, 1.0, 1.0$ \\
  
  \bottomrule
\end{tabular}
\end{sidewaystable}

\end{document}